\documentclass{article}

\usepackage{arxiv}

\usepackage[utf8]{inputenc} 
\usepackage[T1]{fontenc}    
\usepackage{hyperref}       
\usepackage{url}            
\usepackage{booktabs}       
\usepackage{amsfonts}       
\usepackage{nicefrac}       
\usepackage{microtype}      
\usepackage{amsmath}
\usepackage{cleveref}       
\usepackage{lipsum}         
\usepackage{graphicx}
\usepackage{natbib}
\usepackage{doi}
\usepackage{placeins}
\usepackage{framed,multirow}
\usepackage{tabularx}
\usepackage{placeins}

\usepackage{subcaption}

\usepackage{amssymb}
\usepackage{latexsym}
\usepackage{colortbl}

\usepackage{tabularx}

\usepackage{url}
\usepackage{xcolor}
\definecolor{newcolor}{rgb}{.8,.349,.1}
\usepackage{hyperref}

\title{A Multi-Modal Explainability Approach for Human-Aware Robots in Multi-Party Conversation}

\newif\ifuniqueAffiliation
\uniqueAffiliationtrue

\ifuniqueAffiliation 
\author{ \href{https://orcid.org/0000-0002-6396-9770}{\includegraphics[scale=0.06]{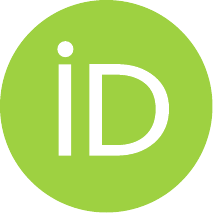}\hspace{1mm}Iveta Bečková$^*$}\\
	Faculty of Mathematics, Physics and Informatics\\
	Comenius University Bratislava\\
	Bratislava, 842 48, Slovak Republic \\
	\texttt{iveta.beckova@fmph.uniba.sk} \\
	\And
	\href{https://orcid.org/0000-0003-3799-7038}{\includegraphics[scale=0.06]{orcid.pdf}\hspace{1mm}Štefan Pócoš$^*$}\\
	Faculty of Mathematics, Physics and Informatics\\
	Comenius University Bratislava\\
	Bratislava, 842 48, Slovak Republic \\
	\texttt{stefan.pocos@fmph.uniba.sk} \\
    \And
    \href{https://orcid.org/0000-0002-6376-9963}{\includegraphics[scale=0.06]{orcid.pdf}\hspace{1mm}Giulia Belgiovine}\\
	CONTACT Unit\\
	Italian Institute of Technology\\
	Genova, 16152, Italy \\
	\texttt{giulia.belgiovine@iit.it} \\
	\And
	\href{https://orcid.org/0000-0003-1719-3745}{\includegraphics[scale=0.06]{orcid.pdf}\hspace{1mm}Marco Matarese}\\
	CONTACT Unit\\
	Italian Institute of Technology\\
	Genova, 16152, Italy \\
	\texttt{marco.matarese@iit.it} \\
    \And
    \href{https://orcid.org/0000-0002-6029-3138}{\includegraphics[scale=0.06]{orcid.pdf}\hspace{1mm}Omar Eldardeer}\\
	CONTACT Unit\\
	Italian Institute of Technology\\
	Genova, 16152, Italy \\
	\texttt{omar.eldardeer@iit.it} \\
    \And
    \href{https://orcid.org/0000-0002-1056-3398}{\includegraphics[scale=0.06]{orcid.pdf}\hspace{1mm}Alessandra Sciutti}\\
	CONTACT Unit\\
	Italian Institute of Technology\\
	Genova, 16152, Italy \\
	\texttt{alessandra.sciutti@iit.it} \\
	\And
	\href{https://orcid.org/0000-0002-9282-9873}{\includegraphics[scale=0.06]{orcid.pdf}\hspace{1mm}Carlo Mazzola}\\
	CONTACT Unit\\
	Italian Institute of Technology\\
	Genova, 16152, Italy \\
	\texttt{carlo.mazzola@iit.it} \\}

\else

\fi

\renewcommand{\undertitle}{Accepted in "Computer Vision and Image Understanding" journal on January 23, 2025}
\renewcommand{\shorttitle}{\scriptsize{A Multi-Modal Explainability Approach for Human-Aware Robots in Multi-Party Conversation}}

\hypersetup{
pdftitle={A_Multi-Modal_Explainability_Approach_for_Human-Aware_Robots_in_Multi-Party_Conversation},
pdfsubject={},
pdfauthor={Iveta Bečková, Štefan Pócoš, Giulia Belgiovine, Marco Matarese, Alessandra Sciutti, Carlo Mazzola},
pdfkeywords={Human Activity Recognition, Explainable AI, Transparency, Attention, Human-Robot Interaction, Addressee Estimation},
}

\begin{document}
\maketitle

\begin{abstract}
The addressee estimation (understanding to whom somebody is talking) is a fundamental task for human activity recognition in multi-party conversation scenarios. Specifically, in the field of human-robot interaction, it becomes even more crucial to enable social robots to participate in such interactive contexts. However, it is usually implemented as a binary classification task, restricting the robot's capability to estimate whether it was addressed or not, which limits its interactive skills. For a social robot to gain the trust of humans, it is also important to manifest a certain level of transparency and explainability. Explainable artificial intelligence thus plays a significant role in the current machine learning applications and models, to provide explanations for their decisions besides excellent performance.
In our work, we a) present an addressee estimation model with improved performance in comparison with the previous state-of-the-art; b) further modify this model to include inherently explainable attention-based segments; c) implement the explainable addressee estimation as part of a modular cognitive architecture for multi-party conversation in an iCub robot; d) validate the real-time performance of the explainable model in multi-party human-robot interaction; e) propose several ways to incorporate explainability and transparency in the aforementioned architecture; and f) perform an online user study to analyze the effect of various explanations on how human participants perceive the robot.
\end{abstract}

\keywords{Human Activity Recognition \and Explainable AI \and Transparency \and Attention \and Human-Robot Interaction \and Addressee Estimation }

\def\thefootnote{*}\footnotetext{These authors contributed equally to this work}\def\thefootnote{\arabic{footnote}}

\section{Introduction}
\label{introduction}

The endeavor to decode human intentions and behavior with advanced computer vision techniques is central to the challenge of developing human-aware technologies that can truly understand and support humans in various everyday scenarios. This is particularly evident in robotics, where the development of embodied agents capable of autonomous and meaningful interaction with humans is facilitated by human activity recognition algorithms, granting them a level of human awareness.
In humans, the recognition of intentions related to social interaction is not limited to high-level reasoning abilities but anchors its roots in the visual system \citep{mcmahon2023seeing}. It follows that visual information is crucial to processing and properly understanding social dynamics, and computer vision models represent an integral component of robots' socio-cognitive abilities.

As the hallmark of human-centered technology, the development of interactive robots is increasingly focused on fostering human trust in artificial systems. Therefore, the accuracy and robustness of the performances are essential but not the only indicators of the system's reliability. Explainability and transparency are two other critical aspects of designing and evaluating reliable interactive robots \citep{wortham2016does}. Both allude to the understandability of the system: transparency predominantly refers to the visibility of underlying processes leading to a reduction of ambiguity regarding a behavior \citep{selkowitz2017displaying}, whereas explainability is related to the capability of a system to exhibit the reasons behind its outputs, decisions or behaviors \citep{ciatto2020agent, miller2019explanation}. 
These two qualities are desirable from a dual perspective. In the eyes of developers, who need deep comprehension of the robot to design and assess its functioning, and from the point of view of users, who should be able to intuitively interact with the artificial system \citep{Sciutti_Humanizing}.  

In the context of Human-Robot Interaction (HRI), communication is a specific type of interaction. Following \cite{JakobsonLinguisticsandPoetics}, communication involves the exchange of messages between an addresser (who sends the message) and an addressee (who is entailed to receive it). In the case of spoken messages, the communication is verbal but usually comprises non-verbal elements such as gaze, gestures, poses, etc., which are grasped via vision and are often necessary to properly contextualize the message and to address it to the correct agent \citep{Skantze2021}. Even though HRI studies rarely go beyond dyadic interactions, the final goal is often bringing robots to social environments, where they are often required to deal with more than one person and, to achieve this aim, be aware of basic social cues ruling multi-party conversations. 

Addressee Estimation (AE), i.e., the ability to understand to whom a speaker is directing their utterance \citep{Skantze2021}, is a specific case of human activity and intention recognition. The speaker identification and correct conversion of speech into text are necessary but not sufficient elements to engage in multi-party conversations. Thus, AE has become a key factor in HRI. As humans, we deeply exploit non-verbal behavior to indicate whom we are addressing \citep{Auer2018gaze, Ishii2016}: an ability that robots could greatly take advantage of to engage in conversations more smoothly. Without it, the understanding of the addressee would exclusively depend on the context of the dialogue or specific keywords (such as the name of the addressee), leading to a loss of fluency in the conversation and an increased likelihood of errors.

Taking into account explainability in the context of human-activity-aware robots requires considering the concept from different perspectives: not only the generation of explanations within the architecture and models controlling the robot's behavior but also their communication to and reception by users. Hence, this work seeks to bind together such diverse points of view that cannot be examined separately.
Starting from this broader approach and with the final aim to endow a social robot with explainable addressee estimation skills for multi-party conversation, we address several key gaps in the field of addressee estimation (AE) for human-robot interaction (HRI). First, we introduce the concept of explainability into AE, leveraging attention-based mechanisms, which, to our knowledge, have not been previously explored in this context. Second, we tackle the limitations of existing AE approaches, which typically only determine if the robot is being addressed, by enhancing the system to identify specific addressees in multi-party conversations. Building on the work of \citet{mazzola}, we optimize their addressee direction estimation model and integrate it into a modular architecture for multi-party interaction, incorporating additional components to enable the robot not only to understand basic communication dynamics but also to leverage communication cues and infer the presence of new people in the environment. Finally, while explainability has been studied in related areas such as sentiment analysis and emotion recognition, we provide the first real-time explainable solution for robotic behavior in multi-party conversations, offering both enhanced transparency and practical utility for HRI scenarios.

The contribution of this work is divided into two intertwined steps, whose methodology is described in Section~\ref{sec:methods}. Specifically, in Subsection~\ref{sec:neural_network}, we design and train an attention-based neural network to optimize a former AE model \citep{mazzola} while extracting explanations at different stages of the inference. In Subsection~\ref{sec:implementation}, we deploy the newly developed explainable model in a modular robotic architecture to enable the iCub robot to engage in multi-party conversation, implement a multi-modal system to provide real-time explanations of its behavior, and explore the users' reception of different modalities of explanation (verbal, embodied, graphical) in a user study.

\section{Related work}
\label{related_work}

\subsection{Attention-based and explainable neural networks} 
Explainable models are becoming increasingly important in machine learning \citep{arrieta_XAI}. Early efforts focused on using simple models to ensure inherent interpretability. However, as deep learning models improved and outperformed simpler ones \citep{krizhevsky_alexnet}, achieving even superhuman performance in some tasks \citep{kaiming_delving}, their lack of transparency became a significant concern. This has made explainability essential, particularly for deploying deep models in critical applications.

Two widely researched and popular methods for deep learning interpretability are SHAP \citep{lundberg_shap} and LIME \citep{ribeiro_lime}. These, along with similar approaches, train a simpler surrogate model to approximate the black-box model's outputs, making it inherently more explainable.

Latent representations of inputs are often analyzed to understand model behavior. However, their high dimensionality often makes visualization and explanation challenging. 
To achieve this, dimensionality-reduction techniques can project the complex data into 1D, 2D, or 3D space, enabling clearer data visualization. Principal Component Analysis (PCA), t-distributed Stochastic Neighbor Embedding (t-SNE), and Uniform Manifold Approximation and Projection (UMAP) are commonly used to reduce the space dimensionality, each having its advantages and disadvantages \citep{Maaten_TSNE, McInnes2018}.

On the other hand, there is often a need to generate precise explanations for the image classification domain. For this, saliency maps, i.e., the importance of image regions for predictions, are often investigated \citep{simonyan_deep, zhou_cam, seljavaru_gradcam}. Using these methods, it is possible to peek inside the black-box model processing an image. A downside of these approaches is their dependence on a specific type of architecture (smooth gradients, convolutions, etc.); otherwise, they are often unable to produce satisfying results.

A branch of research, currently setting the state-of-the-art (SOTA) in the majority of tasks, was initiated by designing the transformer architecture \citep{vaswani_attention} followed by its adaptation for the image domain \citep{dosovitskiy_vit}. Thanks to their attention mechanism, transformers are also often considered more interpretable \citep{kashefi2023explainability}, as one can extract the attention weights during the forward pass. 

Thus, in this work, we leverage the idea of attention in general \citep{shen_dissan, niu_a_review} and the many ways of implementing it in the specific downstream task of AE where, to our knowledge, explainability was never taken into account before.

\subsection{Multi-party conversation in HRI}
The management of multi-party conversations requires robots to be endowed with human activity recognition capabilities. 
Several tasks need to be solved to this aim, not only sound detection and natural language understanding, which are essential to receiving the message. Speaker recognition and diarization, turn-taking, and addressee estimation are crucial problems that need to be tackled to assess beyond the ``what" of the message, the ``who" and the ``to whom" of each utterance \citep{Gu2022}. Endowed with multiple sensors, robots can solve multi-party conversation problems with a multi-modal approach to recognize the scene and human intentions via audio, vision, and interpreting information coming from the conversation context \citep{Bilac2017, Dhaussy2023, Bae2023, Lemon2024}.

Unlike dyadic scenarios, multi-party conversations introduce the challenge of determining who should take the turn when the speaker yields. For the conversation to continue, correctly identifying the addressee(s) typically indicates the next speaker.
Keywords, gaze, pose, para-verbal cues, and contextual information have all proven useful for Addressee Estimation (AE) \citep{Skantze2021}. AE enhances robots' conversational abilities by informing both \textit{when} to intervene and \textit{how} to do so. 

In the majority of cases, works tackling AE during the interaction with artificial agents designed rule-based algorithms \citep{Richter2016}, machine learning models \citep{Turnhout2005, Huang2011, Sheikhi2013}, deep neural networks \citep{mazzola, Tesema2023} or large language model (LLM) based techniques \citep{Lemon2024}, grounded on multiple modalities and features to cope with the ambiguity and unpredictability of human behaviors in real-time interactions. 
However, most current models predict only whether the robot was addressed in a binary way \citep{Turnhout2005, Huang2011, Sheikhi2013, Tesema2023, Lemon2024}. This approach is unsuitable for multi-party interactions; therefore, it is insufficient for the robot to effectively engage in conversations with more than two humans without pre-determined knowledge about other users' presence. 

To resolve this limitation, \cite{mazzola} adopted a deep-learning approach to estimate the direction of the addressee from the robot's perspective, training the model on HRI data collected with a Nao robot \citep{vernissage1}. The same model was then ported and tested with a pilot experiment on the iCub platform \citep{mazzola2024real}, but not yet implemented in an architecture for multi-party conversation. 
After optimizing the approach of \cite{mazzola}, we implemented the new explainable AE model into a more complex modular architecture to allow the robot to identify the addressee outside its field of view.

\subsection{Explainability in HRI}
Explainable Artificial Intelligence (XAI) has predominantly been explored in the human-computer interaction field \citep{lai2021towards, gambino2022considering} but there are still a few studies about XAI within the HRI context \citep{de2017people}. Robots introduce additional layers of complexity to the explainability problem than virtual agents due to their embodiment and the higher number of interaction modalities they can handle \citep{setchi2020explainable}.

Regarding robot planning, \cite{chakraborti2017plan} generated explanations while trying to resolve the discrepancies between the robot and human's internal models. Diversely, \cite{tabrez2019explanation} tackled the problem by focusing on users' task understanding to detect incomplete or incorrect beliefs about the robot's functioning. Instead, concerning human-robot collaboration, \cite{matarese2023ex} focused on explainable robots' influence when providing explanations that consider their common ground, also regarding people's personality traits \citep{matarese2023natural}.
 
Visual explanations have also been explored in robotics \citep{maruyama2022visual, halilovic2023visuo}. Notably, \cite{zhu2022affective} introduced a multi-modal framework combining visual and verbal explanations for facial emotion recognition. Additionally, \cite{sobrin2024enhancing} presented a preliminary study on a vision-language model enabling robots to generate explanations by integrating log and video data.

In recent years, the HRI community has shown a growing interest in verbal explanations, destined to grow further because of the spreading of LLMs. For example, \cite{stange2022self} designed and developed a dialogical model for explanations in HRI, which allows robots to reply to human users' requests and explain their internal state. The authors stressed the iterative nature of their model in managing the explanatory processes as dialogues. Task understanding has also been investigated from a dialogical perspective focusing on the role of negation in human-robot explanatory exchanges \citep{gross2023scaffolding}. Moreover, to allow artificial agents to adapt their explanations to their partners' understanding, \cite{robrecht2023snape} implemented a linguistic explainer model that constructs and employs a partner model.

In the context of multi-party conversation, explainability has been investigated for sentiment analysis of social media dialogues \citep{sinha2021explaining} and emotion recognition with multi-modal attentive learning \citep{arumugam2022multimodal}. However, to our knowledge, there is no approach to provide real-time explainable and transparent solutions for robot behavior in multi-party interaction.

\section{Methods}
\label{sec:methods}

\subsection{Design of the attention-based explainable AE model} 
\label{sec:neural_network}

The development of our explainable AE model consists of two steps: in the first step, we focus on enhancing the classification accuracy with respect to the previous SOTA in the same task \citep{mazzola}, which represents our baseline (see Paragraph~\ref{sec:baseline}). This way, we obtain a first model, which we refer to as the Improved Addressee Estimation (IAE) model. In the second step, we modify this model using inherently explainable modules based on attention (see Paragraph~\ref{sec:attention}) to extract additional information during addressee estimation. We refer to the second model as Explainable Addressee Estimation (XAE) model\footnote{The implementation can be found here: \url{https://github.com/iveta331/Explainable-Addressee-Estimation}}.

\subsubsection{Improved Addressee Estimation model} \label{sec:baseline}

Following \cite{mazzola}, we use the Vernissage dataset \citep{vernissage1, vernissage2} to train an Improved model for the Addressee Estimation task (IAE model). To the best of our knowledge, the Vernissage dataset is one-of-a-kind and designed specifically for solving the task of addressee classification in human-robot interaction.
The dataset contains recordings of multi-party conversations from the robot's point of view. In each conversation, one robot and two human participants are engaging in a conversation about paintings on the wall. The conversations are manually labeled with the relative position of the addressee from the robot's point of view. The possible labels are ROBOT, RIGHT, LEFT, GROUP, and NO-LABEL.

To consistently compare with the baseline \citep{mazzola}, we only use the conversation parts in which ROBOT, RIGHT, or LEFT is the target. We also follow the same data pre-processing and data augmentation. To fully leverage the Vernissage dataset, we further augment the input images using commonly recognized techniques. It involves diversifying the lightning conditions (slight changes in brightness, contrast, saturation, and hue), applying spatial changes (rotation and crop), and blurring the images. The exact ranges for each of the data augmentation methods are detailed in Tab.~\ref{tab:hyperparameters}. The view from the robot's camera is split into two parallel data streams (face images and body-pose vectors), which serve as input to the network and later are merged to form a combined representation.

The model from \citep{mazzola} is further modified by reducing the number of output neurons from the convolutional neural network (CNN) processing the facial information. Also, we replace the convolution on body-pose vectors with fully connected layers. The output dimensionality is chosen to make the length of the outputs from face and pose models similar, allowing for more sophisticated data-fusion methods. Thanks to these modifications, our  IAE model improves the current SOTA while significantly reducing the number of trainable parameters $\approx$135 folds (from 91,706,749 to 677,623).

\subsubsection{Explainable addressee estimation model} \label{sec:attention}
In this section, we describe our neural network architecture (XAE model) that, in addition to yielding accuracy comparable with the IAE model, combines multiple attention-based components, allowing us to extract human-readable explanations. 

\begin{figure}[t]
	\centering
	\includegraphics[width=.65\textwidth]{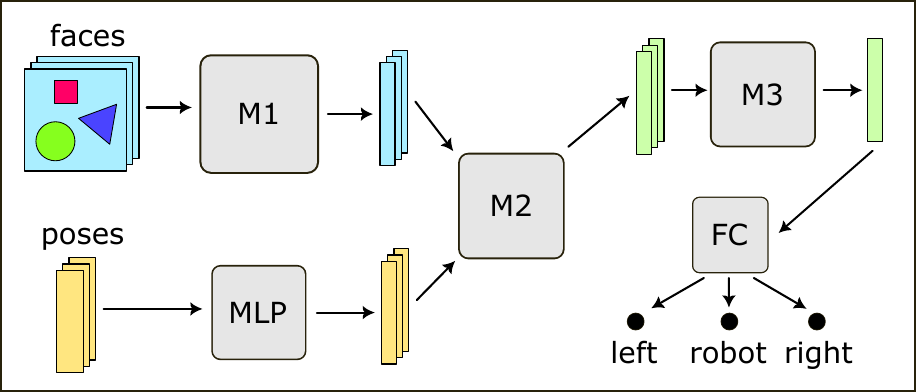}
	\caption{Illustration of the addressee classification workflow. The sequence of faces is embedded using a vision transformer (M1), whereas poses are processed via an MLP. These embeddings are then fused using an intermediate network (M2), and their representation for each time frame is processed by a recurrent network enhanced with attention, forming a unified embedding of the whole utterance. The final step is the mapping to three output options via a fully-connected (FC) layer.}
	\label{fig:general_scheme}
\end{figure}

The information flow in our ``explainable'' architecture is distinct from our IAE model. First, we utilize a vision transformer instead of a convolutional network to obtain the face representation \citep{dosovitskiy_vit}\footnote{For ViT implementation, we use \citep{timm}.}. 
Second, we insert an additional shallow model to fuse the face and pose information. Third, we alter the penultimate processing step using a tailored attention mechanism in the recurrent neural network. This way, we achieve the embedding calculation containing means to provide us with importance scores for each frame. The overall scheme of the addressee estimation is shown in Figure~\ref{fig:general_scheme}.

Most of the above-listed components added to enhance the explainability of the network can be classified as attention mechanisms. A typical approach to incorporating attention into machine learning is to compare similarities of latent representations utilizing the dot product \citep{niu_a_review}. The resulting scores can be used for further activation propagation, where, in theory, only the relevant information is further processed. While it is seemingly straightforward, the information flow and the attention computations can be complex enough to fully replace entire layers of deep networks.

\paragraph{Merging modalities}
After forming the face embedding ($\boldsymbol{f}_t \in \mathbb{R}^{d_{\text{face}}}$) using a vision transformer and a pose embedding ($\boldsymbol{p}_t \in \mathbb{R}^{d_{\text{pose}}}$) using an MLP, we devise a way to combine these representations. The simplest way to achieve this goal is concatenating the two vectors, but it does not admit extracting their relative importance. 

Coherently with our goal, i.e., designing an architecture consisting of multiple components with inherent explainability, we seek to know which modality is more important in each time frame. Inspired by \cite{att_overview}, we opt for the following variant of a scoring function: 
\begin{equation}
    \mathrm{score}(\boldsymbol{v}) = \boldsymbol{w}^T_D\ \mathrm{ReLU}(\boldsymbol{W} \boldsymbol{v} + \boldsymbol{b}),
\end{equation}
where  $\boldsymbol{w}_D\in \mathbb{R}^{d_{\text{inner}}}$, $\boldsymbol{W}\in \mathbb{R}^{d_{\text{inner}} \times d_{v}}$, $\boldsymbol{b}\in \mathbb{R}^{d_{\text{inner}}}$ are trainable parameters, $d_{\text{inner}}$ is a hyper-parameter to be optimized, and $d_v$ is the length of the input vector $\boldsymbol{v}$. Therefore, in this case, $d_{\text{face}}=d_{\text{pose}}=d_v$.

To compare the influence of each of the two vectors (face and pose) at every frame, we calculate their relative contributions $s_{\boldsymbol{f}_t}$ $s_{\boldsymbol{p}_t}$, as:
\begin{equation}
    s_{\boldsymbol{f}_t}, s_{\boldsymbol{p}_t} = \mathrm{softmax}(\mathrm{score}(\boldsymbol{f}_t), \mathrm{score}(\boldsymbol{p}_t)).
\end{equation}
Finally, the single-vector representation of both modalities is formed by an element-wise addition of $\boldsymbol{f}_t$ and $\boldsymbol{p}_t$, using their corresponding weights:
\begin{equation}
    \boldsymbol{r_t} = s_{\boldsymbol{f_t}} \boldsymbol{f_t} + s_{\boldsymbol{p_t}} \boldsymbol{p_t}.
\end{equation}

\paragraph{Recurrent attention}
To form a single vector representation of the whole utterance ($\boldsymbol{r}_1, ..., \boldsymbol{r}_n$), in the baseline architecture, we employ a recurrent network. However, to create the ``explainable'' alternative, we go beyond the ordinary recurrent network and add a form of attention mechanism (as suggested in \cite{att_overview}) that allows us to measure the time frame importance scores while producing the output. Our computation is as follows. 

Let us denote a stacked representation through all the time frames of the embeddings created in the previous step as $\boldsymbol{R} = [\boldsymbol{r}_1, \boldsymbol{r}_2, ..., \boldsymbol{r}_n$]. We linearly project them to produce keys, queries, and values:
\begin{equation}
\boldsymbol{Q}_r = \mathbf{W}^Q \boldsymbol{R},\quad \boldsymbol{K}_r = \mathbf{W}^K \boldsymbol{R},\quad \boldsymbol{v}_r = \mathbf{W}^V \boldsymbol{R}. 
\end{equation}

\begin{figure}[t]
	\centering
	\includegraphics[width=.55\textwidth]{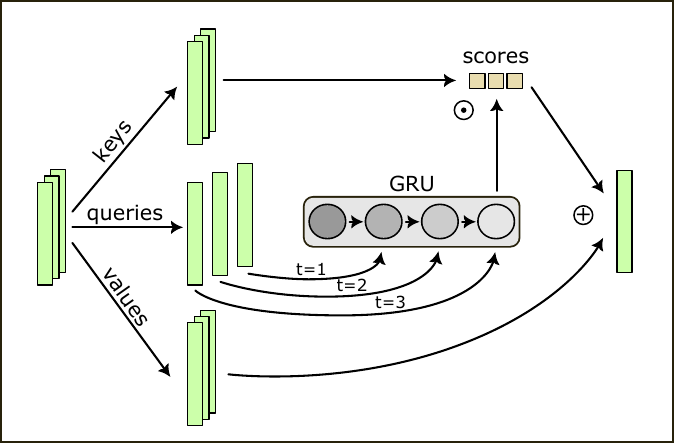}
	\caption{The scheme of the recurrent network augmented with an attention mechanism. The input sequence (left) is projected to form keys, queries, and values. The queries are fed to the GRU network to obtain a single query. That is used to match the keys and compute the scores, which are used to sum up the values (forming the output).}
	\label{fig:attention_gru}
\end{figure}

The queries are then fed one by one into the gated recurrent unit (GRU) network \citep{cho_GRU}, to eventually form the embedding $\boldsymbol{q}$ integrating the information about all the time frames. Using the query embedding $\boldsymbol{q}$, we proceed to the computation of the similarities with keys encoded in the matrix $\boldsymbol{K}$, providing the contribution scores ($\boldsymbol{c}=\boldsymbol{K}\boldsymbol{q}$) of each time frame.
To produce the final utterance representation $\boldsymbol{u}$, we use element-wise addition on elements of $\boldsymbol{V}$, with the weight provided in $\boldsymbol{c}$:
\begin{equation}
    \boldsymbol{u} = \sum_{i=1}^n c_i \boldsymbol{v}_i.
\end{equation}
A fully-connected layer taking $\boldsymbol{u}$ as input produces the final addressee estimation. A graphical illustration of our recurrent block is provided in Figure~\ref{fig:attention_gru}.

\subsubsection{Generating explanations} \label{sec:explanations}
The architectural design of the model is proposed with the intention for the explanations to be inherent. That way, one does not need an extra post-processing step to extract explanations, rendering our model computationally efficient and suitable for use in systems where real-time feedback is necessary. Three types of explanations are included: 1) image saliency, 2) face vs. pose importance, and 3) time frame importance.

\begin{figure}[t]
	\centering
	\includegraphics[width=.70\textwidth]{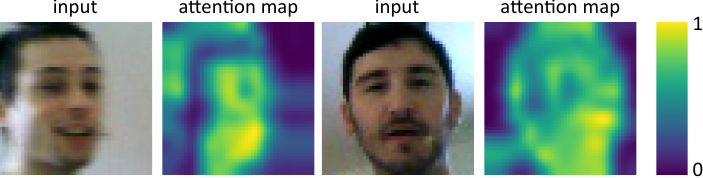}
	\caption{Attention maps extracted from the penultimate layer of the vision transformer employed in the model. The yellow areas indicate active information flow, whereas the blue areas correspond to patches that are not significant at the current layer.}
	\label{fig:attention_face}
\end{figure}

\paragraph{Image saliency} \label{sec:image_saliency}
To process an image via the vision transformer, we first need to split the image into small patches --- non-overlapping squares forming the entire image.
The patches are further embedded and subjected to the attention blocks. Those consist of multiple self-attention layers (multi-head self-attention), residual connections, batch normalization, and fully connected layers, all repeated several times \citep{vaswani_attention}. For visualization purposes, we are primarily interested in the self-attention computation, defined by the formula:
\begin{equation}
    \textrm{Attention}(\boldsymbol{Q}, \boldsymbol{K}, \boldsymbol{V}) = \textrm{softmax}\left(\frac{\boldsymbol{Q}\boldsymbol{K}^T}{\sqrt{d_k}}\right)\boldsymbol{V}.
\end{equation}
The matrices $\boldsymbol{Q}, \boldsymbol{K}$, and $\boldsymbol{V}$ carry the information about each image patch. Thus, visualizing the raw attention scores provided after the softmax computation up-sampled back to the original image size produces maps, highlighting the areas the network uses the most for further processing. A sample depiction of the attention map is provided in Figure~\ref{fig:attention_face}. Since employing multiple parallel heads in the vision transformer is a common practice, we have more visualization options. Because the attention heads extract different features, combining them for visualization can produce more consistent maps.

\paragraph{Face vs. pose} \label{sec:face_vs_pose}
To see the relative importance of face vs. pose information, we can extract the weights used to combine these modalities. Knowing which modality was used more/less can provide the speaker with clues about what was unclear when misclassifications occurred.

Even though the weight extraction is straightforward, some interesting intrinsic properties exist. The general hyper-parameter setup, as well as the whole architecture, have a huge impact on the expressiveness of the model. For a concise analysis of the weights distribution, see Subsection~\ref{sec:AEmodel_analysis}.

\paragraph{Time frame score} \label{sec:time_score}
The attention weights provided in the recurrent network offer an ideal way for us to retrieve information about those time frames, which have the highest impact on the classification. 

Using attention activations, we design a method to automatically generate a verbal cue about the most important part of the prediction. To capture this, we use the average of the attention weights in a sliding window. The average is compared to a threshold $\theta$, and if it crosses the $\frac{1}{k} + \theta$, where $k$ is the length of the sequence, an output sentence is generated based on the sliding window location. For simplicity, in this experiment, we distinguish between three possibilities of the important region: the beginning of the interaction, the middle, and the end. 

\begin{figure}[t]
	\centering
	\includegraphics[width=.75\textwidth]{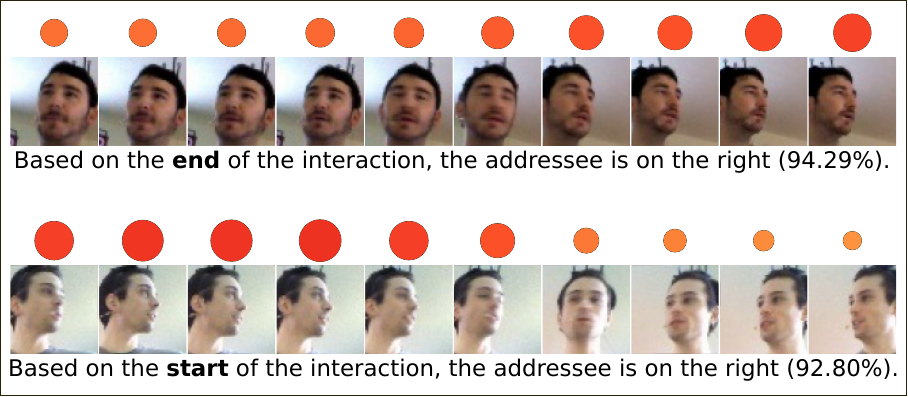}
	\caption{Visualization of two sequences, with their corresponding attention scores (dots) generated by the recurrent network, alongside the generated explanations. The dots' color and size correspond to the attention score magnitude for each time frame.}
	\label{fig:verbal_explanation}
\end{figure}

Two samples of a response, given sequences of length 10, are shown in Figure~\ref{fig:verbal_explanation}. We can see the weights of individual time frames distinguished by the color and size of the red dots. Even though these explanations are particularly easy to interpret, the generation process includes the pose information as well, which is for clarity not included in the image.

\subsubsection{Training and performance evaluation}

To achieve the best possible performance of our models, we perform a hyper-parameter search. We always keep one of the 10 interactions included in the dataset for testing and perform 9-fold cross-validation on the remaining 9 interactions. This cross-validation performance (weighted according to the number of sequences in individual interactions) is being optimized. We use WandB \cite{wandb} with Bayesian search for the optimization.

After choosing all the hyper-parameters, we train a network on nine conversations and test with the remaining one. This is repeated ten times (for all possibilities of the test set). To calculate the final F1 score, we average the results across classes (weighted by the number of samples in the given class) and then across the 10 trials according to the number of sequences in the test sets.

The most extensive hyper-parameter search was performed for the IAE network, as this was the first one being optimized. Values of some of the less important parameters (e.g. parameters of the data augmentation) were then kept the same for all other networks. The list of all optimized hyper-parameters along with all considered values and the chosen values is provided in Table~\ref{tab:hyperparameters}. The hyper-parameters were split into multiple groups, which were being optimized separately. They are ordered and grouped approximately (some parameters were present in multiple groups) in the order in which their value was chosen and fixed. ``Gamma'' controls the learning rate decay. ``Convolutional'' controls the number of output channels: ``small'' = [6, 8, 12, 16], ``medium'' = [8, 12, 16, 32], ``large'' = [8, 16, 32, 64]. The last eight hyper-parameters control additional data augmentation. The final IAE model architecture is summarized in Table~\ref{tab:architecture_IAE}. 

 \begin{table}[!h]
     \centering
     \caption{List of all considered hyper-parameters for the IAE model and their values ([min, max] range for real-valued ones, set of values for discrete). The last column contains the chosen values.}
         \label{tab:hyperparameters}
 	\begin{tabular}{|c|c|c|} 
 		\hline
 		Parameter name & Considered values & Chosen value \\
 		\hline
             \hline
 		num\_epochs & \{5, 6, \dots, 50\} & 15  \\
 		normalisation & \{true\_stats, imagenet\} & true\_stats  \\
 		dropout & \{0, 0.1, 0.2, 0.3\} & 0.2  \\
 		\hline
             act1 & \{ReLU, Tanh\} & Tanh \\
             act2 & \{ReLU, Tanh\} & Tanh \\
             act3 & \{ReLU, Tanh\} & Tanh \\
             hid1 & \{128, 129, \dots, 400\} & 256 \\
             hid2 & \{16, 17, \dots, 40\} & 32 \\
             hid3 & \{16, 17, \dots, 40\} & 32 \\
             out1 & \{10, 11, \dots, 64\} & 32 \\
             out2 & \{10, 11, \dots, 32\} & 20 \\
             out3 & \{8, 9, \dots, 32\} & 20 \\
             optimizer1 & \{SGD, Adam, RMS\} & RMS \\
             optimizer2 & \{SGD, Adam, RMS\} & RMS \\
             optimizer3 & \{SGD, Adam, RMS\} & Adam \\
             post\_fusion & \{LSTM, GRU\} & GRU \\
             convolutional & \{small, medium, large\} & large \\
             \hline
             gamma1 & [0.5, 1] & 0.75\\
             gamma2 & [0.5, 1] & 0.9725\\
             gamma3 & [0.3, 1] & 0.5\\
             learning\_rate1 & [$e^{-10}$, $e^{-7}$] & 0.00018 \\
             learning\_rate2 & [$e^{-10}$, $e^{-2}$] & 0.01 \\
             learning\_rate3 & [$e^{-10}$, $e^{-7}$] & 0.00009 \\
             batch\_size & \{10, 11, \dots , 350\} & 16 \\
             \hline
             brightness & [0, 0.5] & 0.2 \\
             contrast & [0, 0.5] & 0.4 \\
             saturation & [0, 0.5] & 0.45 \\
             hue & [0, 0.25] & 0.135 \\
             \hline
             angle & [0, 45] & 25 \\
             crop & [40, 50] & 50 \\
             kernel\_size & \{1, 3, 5, 7, 9\} & 7 \\
             sigma & [0.1, 3] & 0.8 \\
             \hline
 	\end{tabular}
 \end{table}

 The number of hyper-parameters optimized for the XAE model was much lower. They were optimized in two groups. The first group contained optimization-related parameters, such as gammas and learning rates, the second group contained architecture-related parameters. The resulting values are in Tables~\ref{tab:hyperparams_XAE}~and~\ref{tab:architecture_XAE}, while the latter mentioned also contains information about the network architecture. Activation function in the recurrent network is used on inputs to the GRU unit and outputs from the GRU (i.e., inputs to the final fully-connected layer).

 \begin{table}
     \centering
     \caption{Architecture of our final IAE model. All convolution layers use kernel size 5. Relevant layer-specific hyper-parameters are listed in the brackets.}
     \label{tab:architecture_IAE}
     \begin{tabular}{|c|c|}
         \hline
         \multicolumn{2}{|c|}{Face image model (net 1)}\\  
         \hline
         Input                  & 50x50 px RGB images        \\ 
         \hline
         Convolution            & 8 output channels          \\ 
         Activation (act1)      & Hyperbolic tangent         \\ 
         Convolution            & 16 output channels         \\ 
         Activation (act1)      & Hyperbolic tangent         \\ 
         Max-Pool               & Kernel of size 2$\times$2  \\
         Dropout (dropout)      & p = 0.2                    \\
         Convolution            & 32 output channels         \\
         Activation (act1)      & Hyperbolic tangent         \\
         Convolution            & 64 output channels         \\
         Activation (act1)      & Hyperbolic tangent         \\
         Max-Pool               & Kernel of size 2$\times$2  \\
         Dropout (dropout)      & p = 0.2                    \\
         Flatten                &                            \\
         Fully-connected (hid1) & 256 output neurons         \\
         Activation (act1)      & Hyperbolic tangent         \\
         Fully-connected (out1) & 32 output neurons          \\
         \hline
         \hline
         \multicolumn{2}{|c|}{Pose model (net 2)}     \\
         \hline
         Input                  & Vector of length 54 \\
         Fully-connected (hid2) & 32 output neurons   \\
         Activation (act2)      & Hyperbolic tangent  \\
         Fully-connected (out2) & 20 output neurons   \\

         \hline
         \hline
         \multicolumn{2}{|c|}{Recurrent model (net 3)}      \\
         \hline
         Input                          & Concat. of outputs  \\
         \hline
         GRU block (post\_fusion, hid3) & 32 hidden neurons   \\
         Dropout (dropout)              & p = 0.2             \\
         Fully-connected (out3)         & 20 output neurons   \\
         Activation (act3)              & Hyperbolic tangent  \\
         Fully-connected                & 3 output neurons    \\
         \hline
     \end{tabular}
 \end{table} 

 \begin{table}
     \centering
     \caption{Hyper-parameters of the XAE model. Parameters with indices 1 -- 4 control Vit (M1), MLP for pose vectors, intermediate network (M2), and GRU with attention (M3) respectively.}
     \label{tab:hyperparams_XAE}
     \begin{tabular}{|c|c|}
         \hline
         \multicolumn{2}{|c|}{XAE model hyper-parameters}\\  
         \hline
         gamma1          & 0.70807         \\ 
         gamma2          & 0.96882          \\ 
         gamma3          & 0.51241           \\ 
         gamma4          & 0.9274           \\ 
         learning\_rate1  & 0.00003791         \\ 
         learning\_rate2  & 0.00117708          \\ 
         learning\_rate3  & 0.00185           \\ 
         learning\_rate4  & 0.00006085           \\
         optimizer1      & RMS         \\ 
         optimizer2      & RMS          \\ 
         optimizer3      & RMS           \\ 
         optimizer4      & Adam           \\
         \hline
         batch\_size             & 4         \\
         num\_epochs             & 8         \\
         \hline
     \end{tabular}
 \end{table}

  \begin{table}
     \centering
     \caption{Hyper-parameters/architecture of our final XAE model. For processing face images, we use the standard vision transformer architecture, therefore, for brevity, we only provide the chosen hyper-parameter values. The architecture of the recurrent part is discussed in more detail in Paragraph~\ref{sec:attention}.}
     \label{tab:architecture_XAE}
     \begin{tabular}{|c|c|}
         \hline
         \multicolumn{2}{|c|}{Face image model (ViT)}\\  
         \hline
         Input                        & 50x50 px RGB images\\ 
         \hline
         Patch size                   & 4x4 px         \\ 
         Embedding dimension          & 42          \\ 
         Depth                        & 6           \\ 
         Num. heads                   & 6           \\ 
         Output dimension             & 185         \\
         \hline
         \hline
         \multicolumn{2}{|c|}{Pose model (fully-connected)}     \\
         \hline
         Input                  & Vector of length 54 \\
         Fully-connected        & 73 output neurons   \\
         Activation             & Hyperbolic tangent  \\
         Fully-connected        & 185 output neurons  \\

         \hline
         \hline
         \multicolumn{2}{|c|}{Intermediate (data-fusion) model}      \\
         \hline
         Input                          & Two 185-dimensional vectors  \\
         \hline
         Fully-connected                & 14 output neurons   \\
         Activation                     & ReLU  \\
         Fully-connected                & 1 output neuron   \\
         Normalization                  & Softmax, outputs weights \\
         Weighted sum of inputs         & Output vector of length 185 \\
         
         \hline
         \hline
         \multicolumn{2}{|c|}{GRU with attention (recurrent) model}      \\
         \hline
         Input                          & Sequence of 185-dim. vectors  \\
         \hline
         Values dimension & 81   \\
         Keys dimension & 20   \\
         GRU hidden dim. & 20   \\
         Activation      & Hyperbolic tangent  \\
         \hline
         Fully-connected & 81 inputs, 3 outputs  \\
         \hline
     \end{tabular}
 \end{table} 

Finally, while we utilize the Vernissage dataset for exploring the model's accuracy and the possibilities for generating explanations, to ensure the resulting model is capable of making correct estimates in an online setting, we also retrain our XAE model on both the Vernissage and a dataset collected using the iCub robot \citep{pia_icub_dataset}. The dataset contains five recorded interactions of three people and the iCub, and is labeled into three groups (LEFT, RIGHT, ROBOT), such that it can be used to extend the Vernissage dataset.

\subsection{Implementation of the explainable modular architecture in the robotic platform}
\label{sec:implementation}

After the design, training and testing of the XAE model, the second major contribution of this paper is its deployment in a robotic modular architecture for HRI multi-party conversation, which goes in parallel with an assessment of the two primary features of the architecture: 1) the AE abilities, tested in real-time multi-party interaction, and 2) the explainability and transparency qualities, evaluated through an online user study.

The term artificial cognitive architecture generally refers to computational frameworks that aim to explain and reproduce the fundamental mechanisms of human cognition, such as perception, attention, action selection, memory, learning, reasoning, meta-cognition, or prospective \citep{vernon2014artificial}. 
The development of artificial cognitive architectures able to solve all these tasks with similar performance to humans is a longstanding and unsolved problem \citep{kotseruba202040}. The architecture we propose in this paper is not meant to represent a comprehensive solution for a multi-task robotic architecture. Rather, it is a cognitively-inspired framework supporting the robot's autonomous reasoning, decision-making, and interaction in the context of a multi-party conversation for deploying and evaluating the proposed explainable model in HRI.

\subsubsection{Deployment of the XAE in a modular architecture}
\label{sec:modular_approach}

\begin{figure}[h]
	\centering
	\includegraphics[width=.65\textwidth]{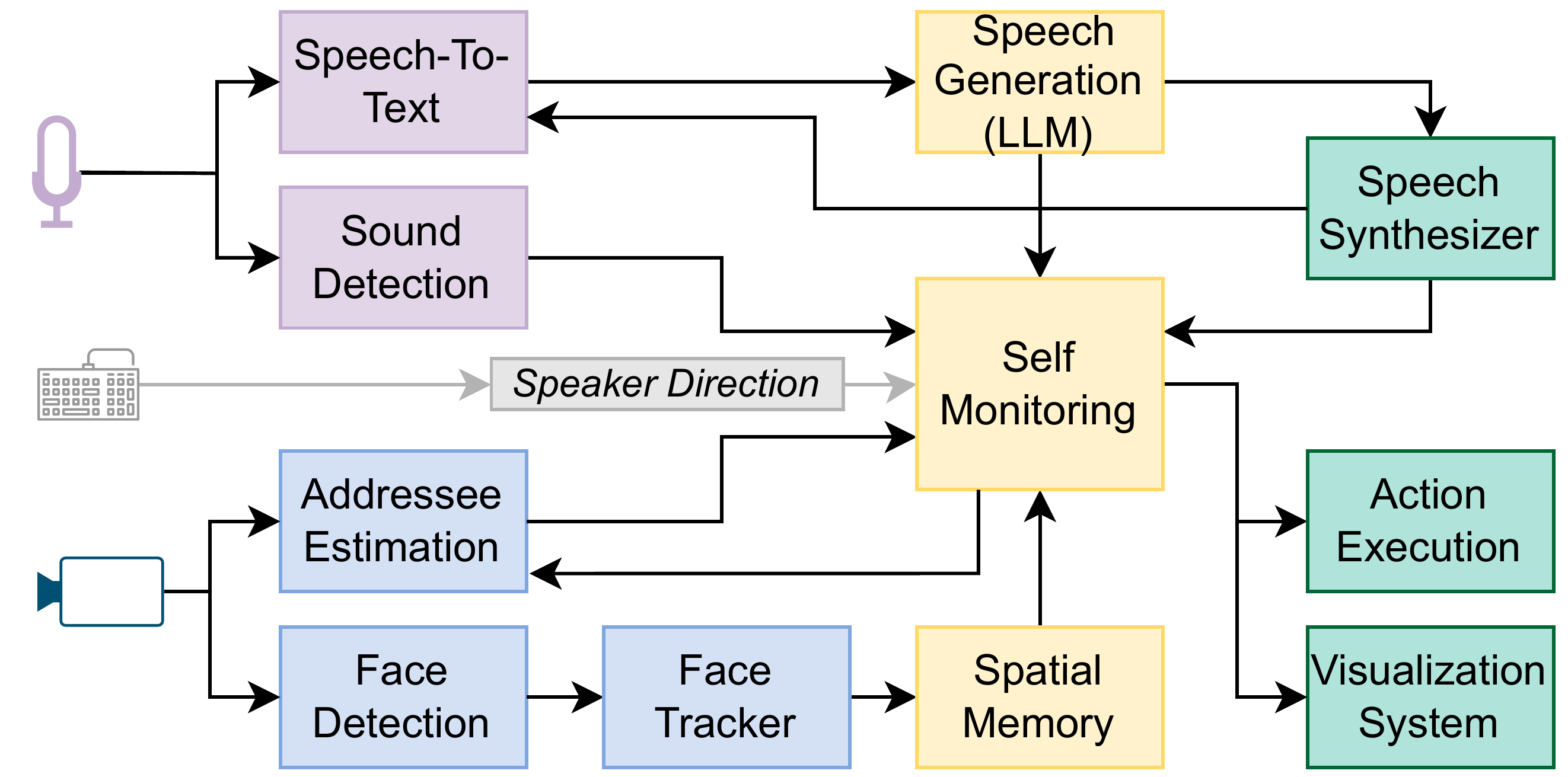}
	\caption{scheme of the modular architecture implemented in the robot iCub. In pink are the audio processing modules, in light blue are visual processing, in yellow are modules to support reasoning processes, and in green are modules to act in the world.}
	\label{fig:cog_architecture}
\end{figure}

Our architecture\footnote{Code available at \url{https://gitlab.iit.it/cognitiveInteraction/explainablemultipartyconversationCVIU.git}} (Figure~\ref{fig:cog_architecture}) is based on the architectural implementation described in \cite{belgiovine2022hri}. It is based on a \textit{modular} approach where each component (i.e., \textit{a module}) implements a specific robot's ability (e.g., detecting faces, processing speech, reasoning about the next best action). Such modules exchange information and communicate with each other through the YARP middleware \citep{metta2006yarp}. To enhance computational resource allocation and optimize time response efficiency, we rely on a distributed approach. This modular setting allows us to incrementally add or update the robot's skills by integrating additional modules into the framework at later stages. 

\paragraph{Audio Perception and Processing}
\label{sec:audio-modules}

Raw audio is given as input to the \textit{Sound Detection} module \citep{Eldardeer2021}.
Based on dynamic thresholding to detect speech and a silence buffer threshold of 1.5 seconds to detect the end of an utterance, this module is programmed to trigger every 0.02 second the \textit{Self-Monitoring} module for the utterance segmentation and the consequent activation of the addressee estimation module.

For the Speech-To-Text module, we use the SOTA Whisper\footnote{\url{https://github.com/openai/whisper}} \citep{radford2022robust} ``small-distil'' model for English language, designed by OpenAI for accurate and robust speech-to-text transcription. Connected in input directly with the microphone and with the speech synthesizer, this module is implemented to synthesize speech while ignoring self-produced utterances.

\paragraph{Vision perception and processing} 

Visual input from the robot's cameras (RGB images of 480x640 resolutions, recorded at 30 fps) is given as input to the \textit{Face Detection} and the \textit{Addressee Estimation} modules to extract higher-level visual features.
The \textit{Face Detection} module extracts the bounding boxes of faces using the Ultralytics YOLOv8 model\footnote{\url{https://github.com/ultralytics/ultralytics}} \citep{Jocher_Ultralytics_YOLO_2023}, which has been adapted for working on the iCub YARP-based framework. 

The \textit{Addressee Estimation} module deploys our XAE model in the YARP middleware. It pre-processes visual information at 12.5 Hz as in \cite{mazzola2024real}, computing the speaker's body pose using a lightweight version of OpenPose \citep{Openpose2019, osokin2018lightweight_openpose} and cropping an image of the speaker's face from the body joints of the head. At each timeframe, the two inputs are given to the XAE model, as described in Section~\ref{sec:neural_network}. After 10 frames, the model returns an estimate of the addressee's direction relative to the speaker and from a first-person perspective: toward left, right, or directly at the robot. Estimates are sent to the \textit{Self-Monitoring} module. The model takes the frames as input cumulatively until the end of the utterance, triggered by the \textit{Sound-Detection} module via the \textit{Self-Monitoring/State Controller} module, when a final estimate is provided.

A \textit{Multiple Objects Tracker} (MOT) is used to generate tracker instances with a unique ID for each bounding box received by the \textit{Face Detector}. Employing a fusion of the Kalman Filter and Hungarian algorithm, this module ensures consistent identity between faces detected in consecutive frames, allowing real-time performances \citep{bewley2016simple}. If necessary, it activates a tracking mode for the robot, enabling it to follow individuals as they move within the environment and update their position when they stop moving \citep{belgiovine2022hri}.

\paragraph{Short-Term Spatial Memory} 

A memory system is a core component of a cognitive architecture to facilitate the accomplishment of high and low-level cognitive abilities such as attention, reasoning, and context understanding. Such abilities become particularly crucial in dynamic interactions involving multiple individuals, enabling the robot to maintain awareness of surrounding events despite its attention being focused on limited evidence.
When developing artificial cognitive architectures, researchers usually adhere to the traditional classification of memory types established in cognitive psychology \citep{ATKINSON196889}, attempting to reproduce their basic functionalities and features.

For the purpose of the current work, our \textit{Spatial Memory} module aims to implement the ability of the robot to remember and update the position of elements of interest (in this case, people) in its surrounding environment with respect to an egocentric frame.
Hence, the module can be viewed as akin to a short-term spatial working memory system, as the information is solely retained to fulfill the current task, namely attending and intervening appropriately in conversations. No information is stored for subsequent learning or retrieval in future interactions.

This module primarily serves to centralize spatial and contextual knowledge.  It aggregates information coming from the MOT and the proprioceptive data derived from the robot's neck and head encoders \citep{roncone2016cartesian}, linking each tracker instance with the robot's head orientation and, after a 3-bin discretization, categorizing it as being to the right, left, or in front of the robot.
This process creates a dictionary associating each individual with their respective robot-centric spatial position. With a frequency of 2 Hz, each element within this dictionary is dynamically updated with properties pertinent to the multi-party conversation task: people's role in the interaction (e.g., speaker, listener, addressee). This dictionary structure works as a knowledge repository, formatted in a readable and standardized manner, which remains readily accessible for online queries by any other module that can efficiently update or retrieve information by using any object or property as a key. For example, the module \textit{Self-Monitoring} can ask the \textit{Spatial Memory} for all the items present to the left of the robot or ask for the location of a specific individual.

\paragraph{Speech Generation} \label{sec:speech_generation}
In order to make the robot able to understand verbal language and give contextual answers, we use an open-weight Large Language Model, specifically the Mistral-7B-Instruct model from MistralAI \citep{jiang2023mistral}, 
and deploy it in the architecture as a YARP-based module. 
Our approach involves simply prompt-engineering specific contextual information to enable the robot to engage meaningfully in multi-party conversations. Details of the prompts used can be found in the \hyperref[app:prompts]{Appendix C}.

\paragraph{Robot Actions and Behaviors} \label{sec:robot-actions-module}
For tracking faces and directing the robot's head towards the identified speaker, we employ the iKinGazeCtrl module -- a controller designed for iCub's gaze, capable of independently steering the neck and eyes  \citep{roncone2016cartesian} toward a point in the cartesian space. Additionally, our action module provides functionalities for controlling facial expressions (by activating LEDs) and synthesizing speech (by using Acapela Text-To-Speech\footnote{\url{https://www.acapela-group.com/}} with the child-like English voice ``Josh'').

\paragraph{Self-Monitoring/State Controller}
This module acts both as Self-Monitoring and as State Controller. As a state machine, it is responsible for handling information coming in inputs and controlling the activation of other modules subsequently. For instance, it takes in input information coming from the \textit{Sound Detection} module to manage the activation of the \textit{Addressee Estimation} based on the start and end of the speaker's utterance.

In parallel, it also monitors all the active modules of the architecture, providing such information to the \textit{Visualization System} to be displayed on the tv-screen. 

\begin{figure}[t]
    \centering
    \includegraphics[width=0.9\textwidth]{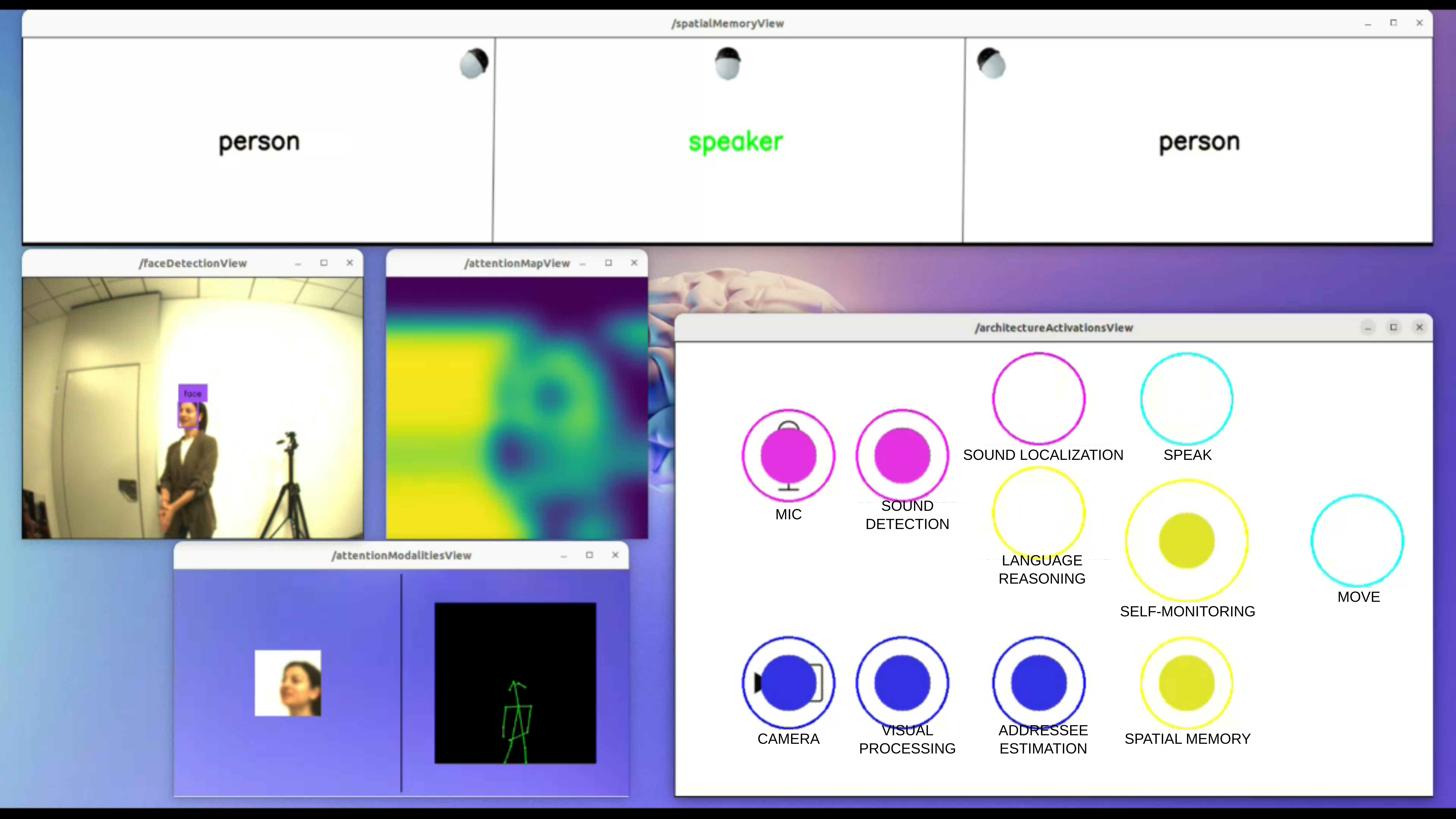}
    \caption{Screenshot of the visual explainability and 
    transparency techniques that were shown on the screen behind the robot during the multi-party conversations. From top to bottom and from left to right: 1. visualization of Spatial Memory module; 2. camera stream with Face Detection information; 3. Attention Map showing the saliency of the input image of the XAE model; 4. the modular architecture with current activations displayed; 5. the relative contribution of the two inputs of the XAE model.}
    \label{fig:screenshot}
\end{figure}

\paragraph{Visualization System} 
An additional module of the architecture is the real-time visualization of the main processes involved in the proposed framework. This includes details about the robot's current state, such as the activation of the modular architecture, which illustrates the ongoing processes supporting the robot's cognitive abilities. Additionally, it showcases the explainable and transparent solutions integrated into the XAE model.
A 50-inch TV screen positioned behind the robot serves as a presentation board for various visualization windows (Figure~\ref{fig:screenshot}).

\begin{figure*}[!ht]
    \centering
    \includegraphics[width=0.9\textwidth]{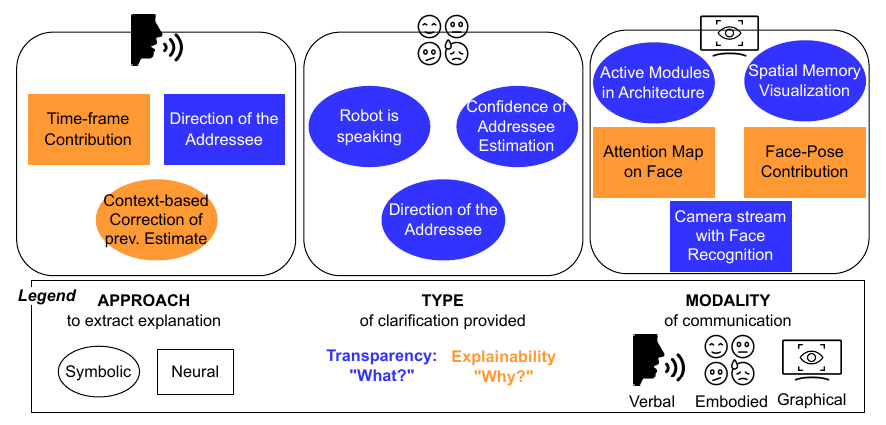}
    \caption{Illustration of our XAI framework. The explainability and transparency techniques we used in our system are ordered by approach (Symbolic or Neural), type of clarification (Transparency or Explainability), and modality of communication (Verbal, Embodied, Graphical).}
    \small\textsuperscript{Icons' attribution goes to Flaticon.com. Specifically, Verbal: Freepik; Emoticons: Khoirul Huda; Screen: Dmytro Vyshnevskyi; Eye inside the screen: Freepik}
    \label{fig:framework}
\end{figure*}

\subsubsection{Evaluation of XAE model performance in real-time multi-party HRI.}
\label{sec:methods:HRIstudy}

In order to check the performance of the XAE deployed in real-time and, therefore, the actual AE capabilities of the above-mentioned architecture during multi-party HRI, we conducted an HRI evaluation study consisting of multiple sessions of interactions between three agents: the iCub robot and two humans. In each session, two researchers impersonated the characters of a fictional scenario in which a service robot (the iCub) was present (home, shopping mall, restaurant). 8 researchers (50\% females) took part in the study, some of them acting in more than one session. We recorded a total of 122 utterances.

The robot's behavior was managed by the architecture described in Section \ref{sec:modular_approach}. The ground truth of the addressee was logged during the experiment by an operator who observed the interaction. For each utterance, the operator logged the addressee (robot or right or left) using the buttons of a GUI designed for this scope.

To evaluate the AE performance of the architecture, for each utterance we collected data coming from different modules: \textit{Sound Detection}, \textit{Sound Localization}, \textit{Addressee Estimation} \textit{Speech-To-Text}, and \textit{Speech Generation}. After the experiments, data were automatically assembled together with the ground truth by using information from the system timestamps. 

To overcome unbalance in class representation (the class ``robot'' was represented 64.75\% of the utterances, whereas ``right'' and ``left'' 18.86\% and 16.39\% respectively), we used a weighted F1 score as a performance indicator. We evaluated the recorded experimental data about the addressee estimation at the different stages of the architecture. Firstly, we evaluated every output of the XAE model considering all its predictions separately (the XAE is programmed to provide an output every 0.8 seconds. Secondly, we evaluated only the final prediction of each utterance, as this is the prediction the robot uses to determine its behavior. Lastly, we recalculated the AE performance, taking into account corrections made by the Spatial Memory (e.g., if the prediction was 'left' but no one was found on the left, the robot considered itself the addressee). 

We approached AE primarily as a Computer Vision problem, although, as humans, we also rely on syntactic and contextual clues within the message to infer the recipient. While we intend to explore the integration of both visual and semantic information to enhance the robustness of the AE model in future work, we analyzed whether, and with what performance, AE could be solved solely as a linguistic problem. 
We adopted a binary approach to determine whether the addressee of each sentence was either ROBOT or NOT-ROBOT. We evaluated three different models: Llama3 by Meta, GPT-3.5-Turbo by OpenAI, and Mistral7b by Mistral AI. Each model was tested with sentences from the conversation of the evaluation study, using both the context prompt employed in the actual interactions and an analysis prompt (refer to ``LLM-based AE Predictions Context'' in the \hyperref[app:prompts]{Appendix C}).
To ensure a fair comparison with the vision-based system, we did not provide the LLM with the chat history, ensuring that the final decision was based solely on the current sentence and the context prompt.
Given the inherent randomness in language models' responses, we considered the average response over 10 iterations for each sentence. Additionally, we repeated the same analysis excluding sentences with explicit references to the addressee's name (e.g., ``Hello Mike, how are you?" or ``Hi, iCub, can you help me with this?"). As a baseline for the Vision approach, we aggregated the left and right classes as NOT-ROBOT and we computed a weighted F1-score on the same binary classification as the LLMs' task. 

\subsubsection{Development of a multi-modal Explainability and Transparency System}
\label{sec:multimodal_system}
To implement a transparent and explainable modular architecture, we developed a framework providing several real-time clarifications about the processes and functionalities of the robot. To this aim, we use diverse techniques (see Figure~\ref{fig:framework}).

Specifically, our solutions are the result of two possible \textit{approaches to extract explanations} \citep{Kerzel2022}: neural or symbolic. The former approach leverages neural network outputs to extract information, whereas the latter represents the information about the system's behavior with a symbolic code specified by the developer.

Another important difference is related to the \textit{type of clarification provided}, which is rooted in how some literature differentiates between the concepts of explainability and transparency \citep{ciatto2020agent, miller2019explanation}. Several accounts we implemented provide the reason behind the decision of the robot (explainability), whereas others describe the current functioning of the robot (transparency). The former answer the question ``why is it happening?'', the latter reply to the question ``what is happening?''. 
Following this classification, we implemented four explainability and seven transparency solutions for the human-robot interaction phase. 

Eventually, several \textit{modalities of communication} are exploited: verbal, embodied, and graphical. 

Figure~\ref{fig:screenshot} shows a screenshot of the five techniques implemented via the graphical modality and displayed in real-time on the screen behind the robot. Two out of these five represent functional aspects of the architecture, i.e., how the architecture is working in terms of current activations and current information in memory. The Spatial Memory provides a 3-bin scheme with instances of people perceived and remembered by the robot, their (robot-centric) position, and conversational role. An additional solution shows the scheme of the robot's modular architecture onscreen, with real-time information about the current activations for each robot's ability. 

The other three graphical representations are implemented as streams of features the robot is paying attention to. The visual streams show, therefore: 
\begin{itemize}
\setlength\itemsep{0em}
  \item the processed output of the iCub's right camera with bounding box information of detected faces;
  \item the attention map of the speaker's face providing information about the salient part of the input image processed by the XAE model;
  \item the two inputs of the XAE model (the speaker's body pose and face image) displayed with size changing in real-time proportionally to the relative importance of their contribution to the final output.
\end{itemize}

Other explainability and transparency techniques designed to clarify the addressee estimation process were implemented 
via the verbal modality. For instance, at the end of other speakers' utterances, the robot provides an explanation for its final estimate of the addressee, saying that its decision relied on the speaker's non-verbal behavior and, more specifically, which part of the speaker's utterance (beginning, end, or the middle) impacted most on the final estimate. 
This enhances the transparency of its subsequent action (turning toward the addressee vs replying to the speaker). Moreover, the robot can correct any evaluation errors made by the XAE model after exploring the environment (e.g., if there is nobody in the estimated direction), and verbally clarify the correction.

Finally, we designed specific actions for the iCub's facial expression to increase the understandability of the robot's behavior via an embodied modality.
iCub is endowed with face LEDs (eyebrows and mouth) that can change color and position and are synchronized with the speech synthesizer.
We specified the iCub mouth (happy or neutral) to be coherent with the confidence of the estimate (happy for confidence over 80\%, neutral otherwise). The color of the LEDs indicates the conversational role of the robot, consistent with the colors displayed in the spatial memory (green for ``speaker'', red for ``addressee'', and white otherwise). The movement of the eyebrows is meant to suggest the direction "left" or "right" of the estimated addressee (see video for clarification). 

\subsection{User Evaluation of the Explainability and Transparency System: Online User Study.} 
\label{sec:user_study}

After developing the XAE model and integrating it into the robotic architecture, we ran a user study to assess external users' perception of the Explainability and Transparency System described in the last Section.

\paragraph{Video Recordings} \label{sec:video_recordings}
We recorded a video of multi-party conversations with the robot\footnote{The entire video can be found at the following link \url{https://www.youtube.com/watch?v=zZF-L0gtRu4}} and uploaded it on the soSci Survey\footnote{\url{https://www.soscisurvey.de/}} platform for questionnaires delivering. 
In the video, the experimenters stage 3 different conversational scenarios, namely interacting with a social robot assistant in a shopping mall, a restaurant, and a domestic setting.  Upon receiving the speaker's position and gazing toward them, the robot utilizes the XAE model to estimate the intended addressee. In the case the estimated addressee is the robot, it replies to the speaker using the Speech Generation module. To ensure meaningful dialogues, we provided the LLM with context prompts tailored to each scene. 

\begin{figure}[t]
    \centering
    \includegraphics[width=.85\textwidth]{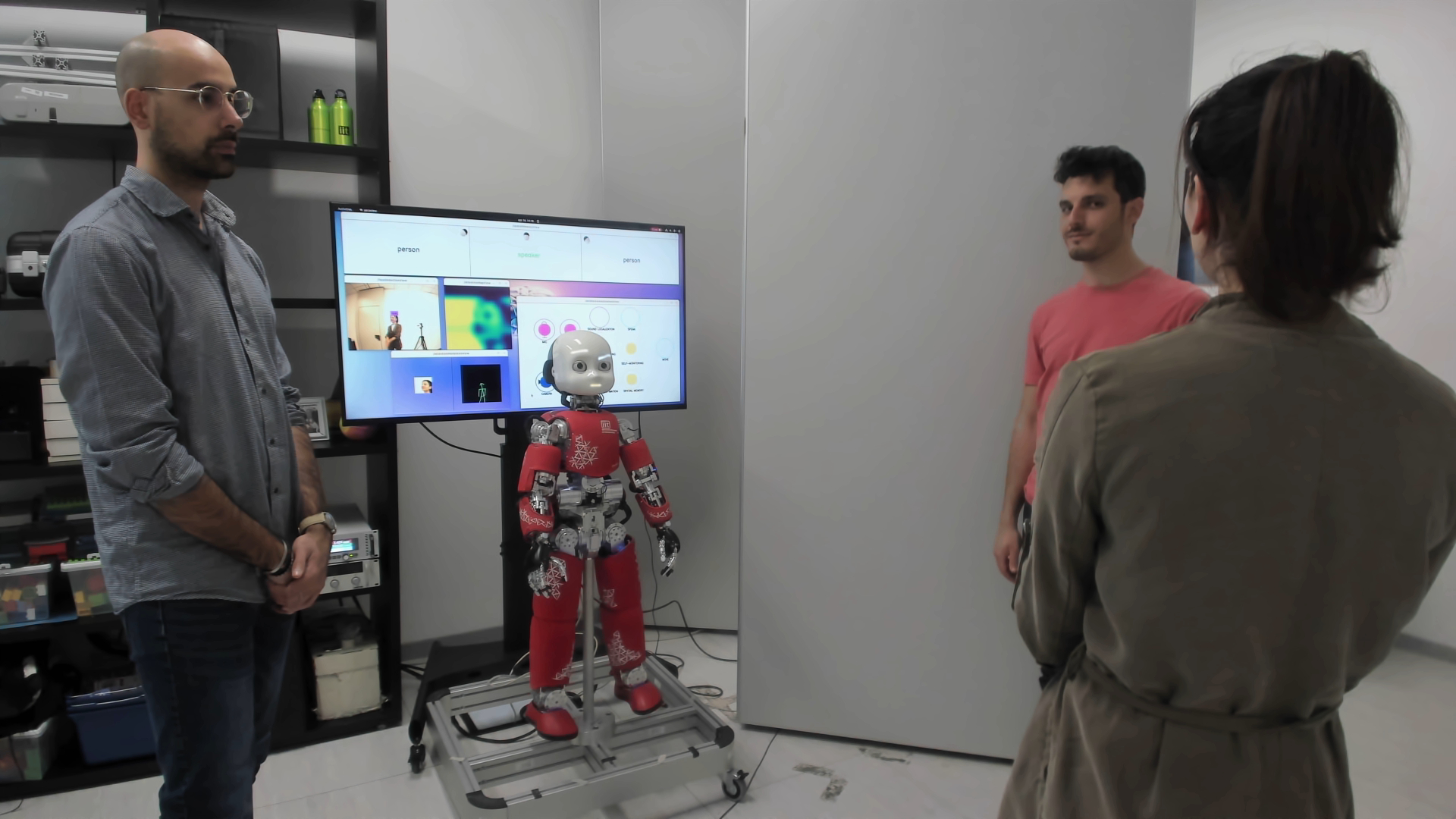}
    \caption{A frame of the multi-party conversation between the robot and the three actors. iCub is expressing a doubting face because it is estimating the addressee of the speaker in front of it; its LEDs are white because it has no role in the current dialogue. Behind the robot, we can see the screen displaying the graphical explainability and transparency techniques (Figure~\ref{fig:screenshot}).}
    \label{fig:interaction}
\end{figure}

Participants were also given an extra introductory video clip to familiarize them with the visual solutions of the Explainability and Transparency System.
The video shows the conversational interactions from an external perspective, with the robot in the middle (Figure~\ref{fig:interaction}). To increase the visibility of important features, we incorporated a zoomed-in view of the robot's face in the bottom left corner, along with the robot's visual explanation system in the bottom right corner of the screen (Figure~\ref{fig:screenshot}).  Participants were instructed to focus on both the overall scene and the robot's explanations.

\paragraph{Demographics}

We performed an online user study with 60 participants (76\% female) with naive users. We designed the survey on soSci\footnote{www.soscisurvey.de} and administered it through Prolific. Participants received a compensation of £9.00 per hour. All participants gave informed consent about collecting and analyzing data anonymously. The study was approved by Comitato Etico Regione Liguria.

\paragraph{Questionnaires} \label{par:Questionnaires}
After viewing the videos, participants were given 7-point Likert scale questionnaires about their perceptions of the robot, and their impressions about the clarifications provided by the implemented Explainability and Transparency System. For the robot's perception, we used items regarding its Warmth and Competence \citep{fiske2007universal}, Likeability (adapted from \cite{spaccatini2019sexualized}), Experience and Agency \citep{hray2007dimensions}, and Cognitive and Affective Trust \citep{bernotat2021fe}. Moreover, we asked them to rate their Satisfaction with the Explanations \citep{hoffman2018metrics}, and their perceived Usefulness and Intrusiveness \citep{conati2021toward}. We asked the participants to individually evaluate the various explanation modalities (verbal, embodied, and graphical), also with open-ended questions,  to gain a complete understanding of their perception of each. 

\paragraph{Statistical Analysis} \label{par:Analysis}

Data obtained from questionnaires have been analyzed in Jamovi Software v. 2.4.11 \citep{jamovi2023} using Mixed Models \citep{gallucci2019gamlj}, Spearman Rank Correlation, and Conditional Mediation analysis \citep{gallucci2020}.

Each of the three XAI parameters (Satisfaction, Usefulness, and Intrusiveness) were employed as the dependent variable in a Mixed Model that computed statistical differences between the three explanation modalities (factor). Participants' ID was applied as a random effect to adjust for each participant's baseline and model the intra-subject correlation of repeated measurements. Bonferroni correction has been applied to assess post-hoc pairwise comparisons between the explanation modalities.

Moreover, we conducted Spearman rank correlation tests to investigate the relationship between the three parameters used to assess participants' evaluation of explanation modalities from the user perspective. For this test, we analyzed the three explanation modalities independently. 

To assess the extent to which the robot's explanations of its behavior contributed to building participants' trust, we examined the relationship between the usefulness of these explanations and the two components of trust: affective and cognitive. To this aim, we performed two tests using Conditional Mediation analysis, one for each component of trust. Affective Trust and Cognitive Trust were used as the dependent variable of the analysis, while Usefulness acted as the covariate. We tested such relationships using the robot's qualities evaluated by participants in the survey as Mediators (Perceived Warmth, Experience, Agency, Likeability, and Competence). In addition, we employed the explanation modality as a Moderation Factor to assess the extent to which each of them contributed to the building of trust.

\section{Results}

\subsection{Evaluation of the AE models.}
\label{sec:AEmodel_performance}

\subsubsection{Performances of the improved AE and XAE models and comparison with SOTA baseline}

\begin{figure}[t]
	\centering
	\includegraphics[width=.69\textwidth]{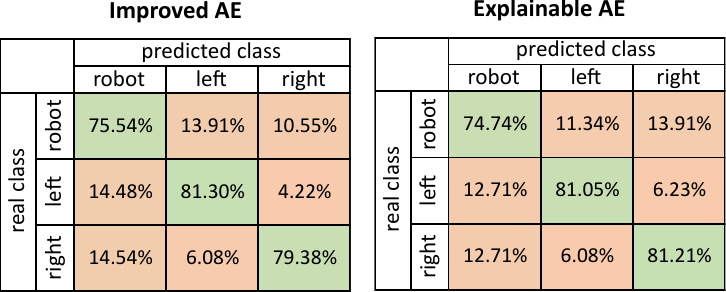}
	\caption{Confusion matrices for the IAE (left) and the XAE (right) model.}
	\label{fig:confusion_matrices}
\end{figure}

To provide a statistically robust evaluation of our proposed models (IAE model as well as the XAE model), the training is repeated five times in total for each test set, using different random seeds. The two models achieve comparable accuracy. The IAE model reaches 79.51\% average F1 score with a standard deviation of 0.56\%, whereas the F1 score for the XAE model is 79.40\% with a standard deviation of 1.06\%. Thus, both models surpass the previous SOTA F1 score of 75.01\%, described in \cite{mazzola}, by $\approx 4.45$\%. When looking at their confusion matrices (Figure~\ref{fig:confusion_matrices}), we see that the models have roughly equal distribution of each type of misclassification. We also see that the models are slightly weaker on recognizing the cases when the addressee is the robot. 

To test our XAE model's capabilities on recordings captured using the iCub, we first analyze its accuracy on the data provided in \cite{pia_icub_dataset} (further referred to as the iCub data). We trained the XAE model on the Vernissage corpus and tested it on the iCub data, with resulting testing accuracy of 66.31\%. This informs us that even though the data distribution is quite different from the Vernissage dataset, the model can successfully estimate the correct addressee most of the time. However, the performance is still notably worse than on the Vernissage data. Thus, to ensure the best possible accuracy of the model when deployed in the iCub, the final model used in the user study was trained on both of the datasets.

\subsubsection{Ablation studies}

As there can be three different attention modules in our network, XAE and IAE can be seen as two possible extremes -- all attention modules on, or all off. However, we can treat them independently, getting six new combinations, and, therefore, six more networks to analyze. By doing this, we get more insight into the contributions of individual attention modules. 

For each of the additional networks, we performed a hyper-parameter search similar to the one done with XAE. All the searched parameters and the optimal values are displayed in Table~\ref{tab:ablations_XAE}, along with F1 scores and standard deviations, computed across 5 different random initializations.

As we can see, all the analyzed networks achieved very similar F1 scores, comparable with the performance of IAE and XAE models. However, the hyper-parameter values differ greatly, meaning that each of the networks can perform well, if a proper hyper-parameter search is conducted. Otherwise, the performance may suffer.

\begin{table*}[t]
  \centering
  \small
     \caption{Optimal hyper-parameter values and resulting performance of networks analyzed during the ablation studies. First three columns describe the networks.}
     \label{tab:ablations_XAE}
     \begin{tabular}{|c|c|c||c|c|c|c|c|c|c|c|c|c||c|c|}
         \hline
         \multicolumn{3}{|c||}{Attention modules} & \multicolumn{10}{c||}{Hyper-parameters} & F1 & std \\  
         \hline
         M1 & M2 & M3 & act3 & act4 & e-dim & out1 & out2 & out3 & opt3 & dim\_qk & dim\_in & dim\_v & & \\ 
         \hline
         \hline
         off & off & on  & - & Tanh & - & 37 & 23 & 20 & - & - & - & 53 & 79.8 & 0.787 \\ 
         off & on  & off & ReLU & - & - & 158 & out1 & 58 & Adam & 83 & 116 & - & 80.0 & 0.643 \\ 
         off & on  & on  & Tanh & Tanh & - & 152 & out1 & 53 & SGD & 90 & 129 & 63 & 81.5 & 0.428 \\ 
         on  & off & off & - & - & 30 & 75 & 19 & 59 & - & - & - & - & 80.5 & 0.609 \\ 
         on  & off & on  & - & Tanh & 180 & 16 & 14 & 22 & - & - & - & 102 & 78.4 & 1.414 \\ 
         on  & on  & off & ReLU & - & 66 & 13 & out1 & 40 & SGD & 167 & 24 & - & 79.1 & 0.765 \\ 
         \hline
     \end{tabular}
  
\end{table*}

\subsection{Quantitative assessment of Explainability in XAE model}
\label{sec:AEmodel_analysis}

When analyzing the distribution of the face vs. pose attention weights, we found that the face and pose information is almost always equally important, i.e., the average score is 0.5. Our resulting XAE model achieves an average score of $0.5$ and deviation of $0.04$. On the other hand, we notice a negative correlation of -0.87 (p = 0.001) between the dimensionality we use for the face and pose embeddings and the logarithm of the importance deviations. The higher the dimensionality, the lower the deviation of attention scores (the lower the distinction rate between the two modalities). Thus, during the optimization process, it is not enough to optimize the model only for the performance, but also for the expressiveness of the explanations, like the deviation of importance scores in this case.

Our experiments also showed that in some cases of networks with lower dimensionalities of the face and pose embeddings, the explanation capability seemed to increase, and the pose information was slightly prevalent.

To further explore the properties of our XAE model in greater detail, we analyze time frame scores with 10-frame-long sequences of the Vernissage dataset. A way to look at the activations is to compute their distribution. When considering only the stack of values independently, they precisely follow a Gaussian distribution with a mean of $0.1$ and standard deviation $0.0055$. Our explanation of the low deviation is that since there are only 10 frames in each sequence, they do not capture a long period of time; thus, their embeddings are usually quite similar. We also observe that the weights change more in the case of frames with greater variability.

\begin{figure}[t]
	\centering
	\includegraphics[width=.79\textwidth]{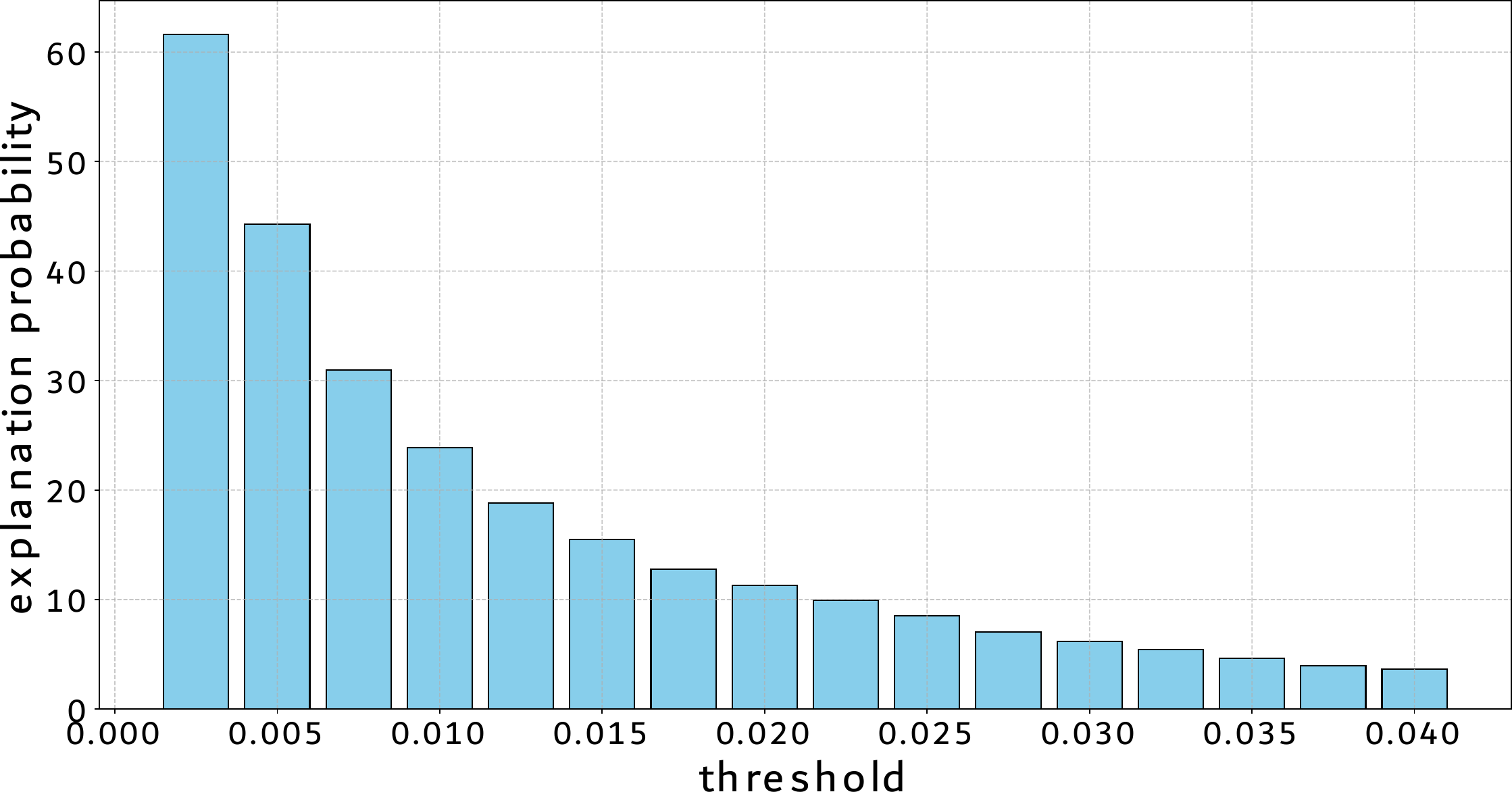}
	\caption{The influence of threshold value ($x$-axis) on the probability ($y$-axis) of a verbal explanation being triggered at the end of a sequence.}
	\label{fig:explain_prob}
\end{figure}

Next, we analyzed the threshold value for triggering an explanation at the end of a sequence. The threshold clearly influences the rate at which the verbal cue is generated. In Figure~\ref{fig:explain_prob} we can see the probabilities of triggering a verbal response for differing threshold values. We empirically verified that threshold values higher than 0.02 yield verbal cues aligning with human expectations. In contrast, by using lower values, the noise patterns sometimes overrule the useful information, yielding responses that are difficult to verify. 

\subsubsection{Comparison of ViT attention maps with Grad-CAM}
Even though using our XAE model we can extract the inherent attention maps of the vision transformer to produce explanations, there are means to generate similar saliency maps for other model families, such as CNNs, present in the baseline model.
One such method is the Gradient-weighted Class Activation Mapping (Grad-CAM) proposed by \citep{seljavaru_gradcam}, which computes the weighted activations, usually of the last convolutional layer, upsampled to the input space.

\begin{figure}[t]
	\centering
	\includegraphics[width=.65\textwidth]{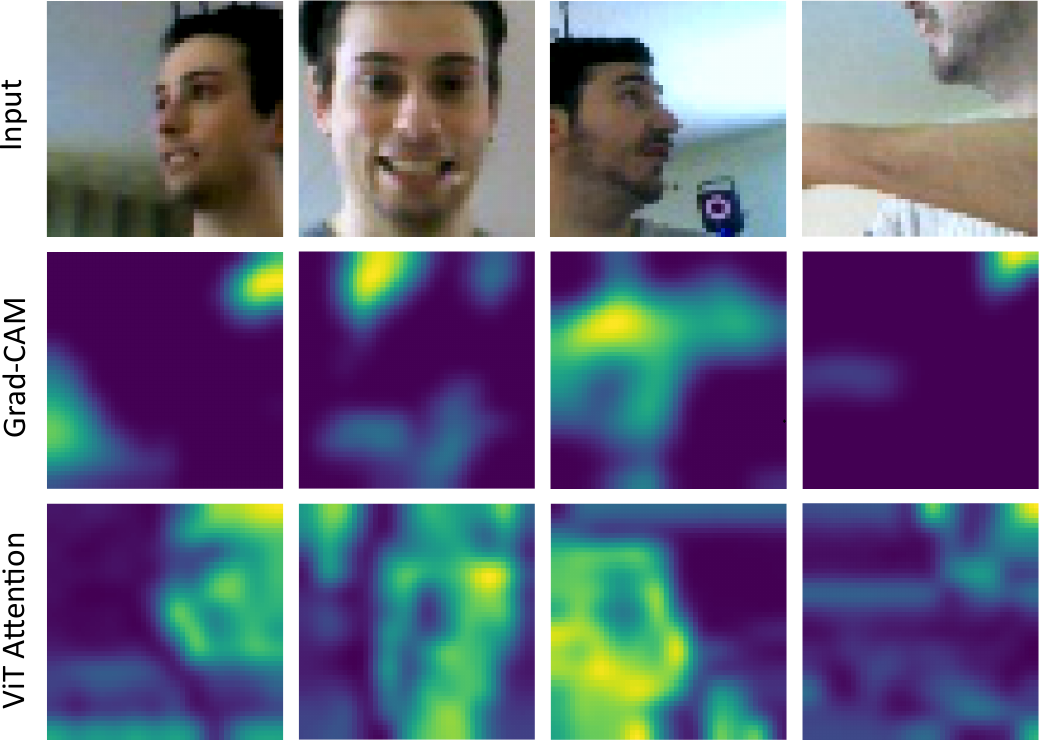}
	\caption{Comparison of saliency maps in IAE and XAE model. In the former, we employ Grad-CAM to approximate the importance of image regions, whereas in the latter, the important image parts are taken from the attention mechanism present in the ViT.}
	\label{fig:grad_cam_comparison}
\end{figure}

In Figure ~\ref{fig:grad_cam_comparison}, we see four examples of the importance of face in the baseline model and the explainable model. In general, the saliency maps of the XAE model are visually more appealing than those produced using Grad-CAM (in the IAE model), yet it is important to consider all factors influencing the quality of the maps. Each method has its own set of hyperparameters, and lacking a fool-proof numerical evaluation of the explanation quality, it can be difficult to pick the best ones.

\subsubsection{Face vs. pose expressiveness in IAE and XAE models}
To further compare the properties of the XAE model with the IAE model, we analyzed the high-dimensional embeddings of face, pose, and their combination (in the case of IAE, their concatenation), using Uniform Manifold Approximation and Projection (UMAP) \citep{McInnes2018}.

\begin{figure}[t]
    \centering
    \begin{subfigure}[t]{0.15\textwidth}
        \centering
        \includegraphics[width=.98\textwidth]{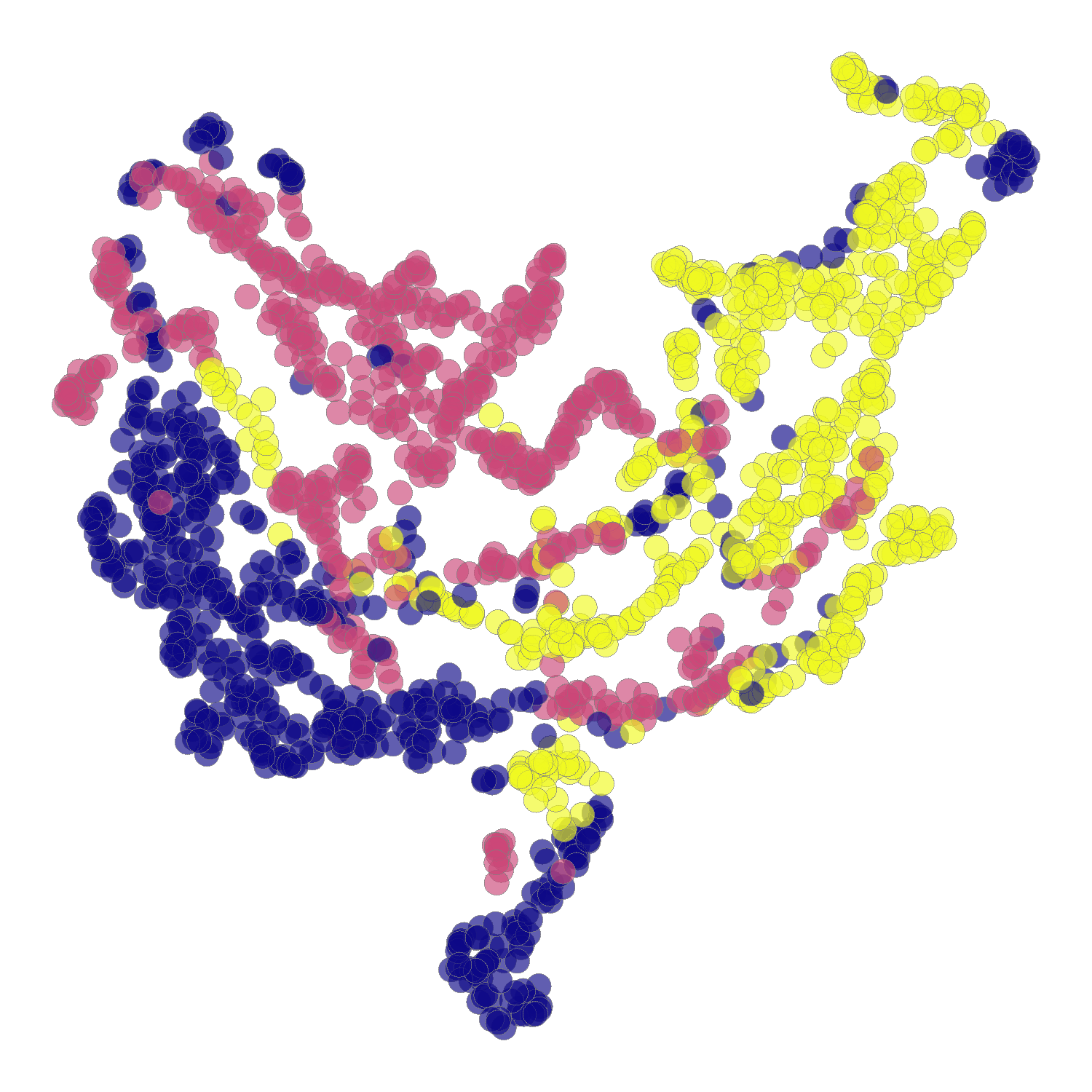}
        \caption{IAE face}
    \end{subfigure}%
    ~ 
    \begin{subfigure}[t]{0.15\textwidth}
        \centering
        \includegraphics[width=.98\textwidth]{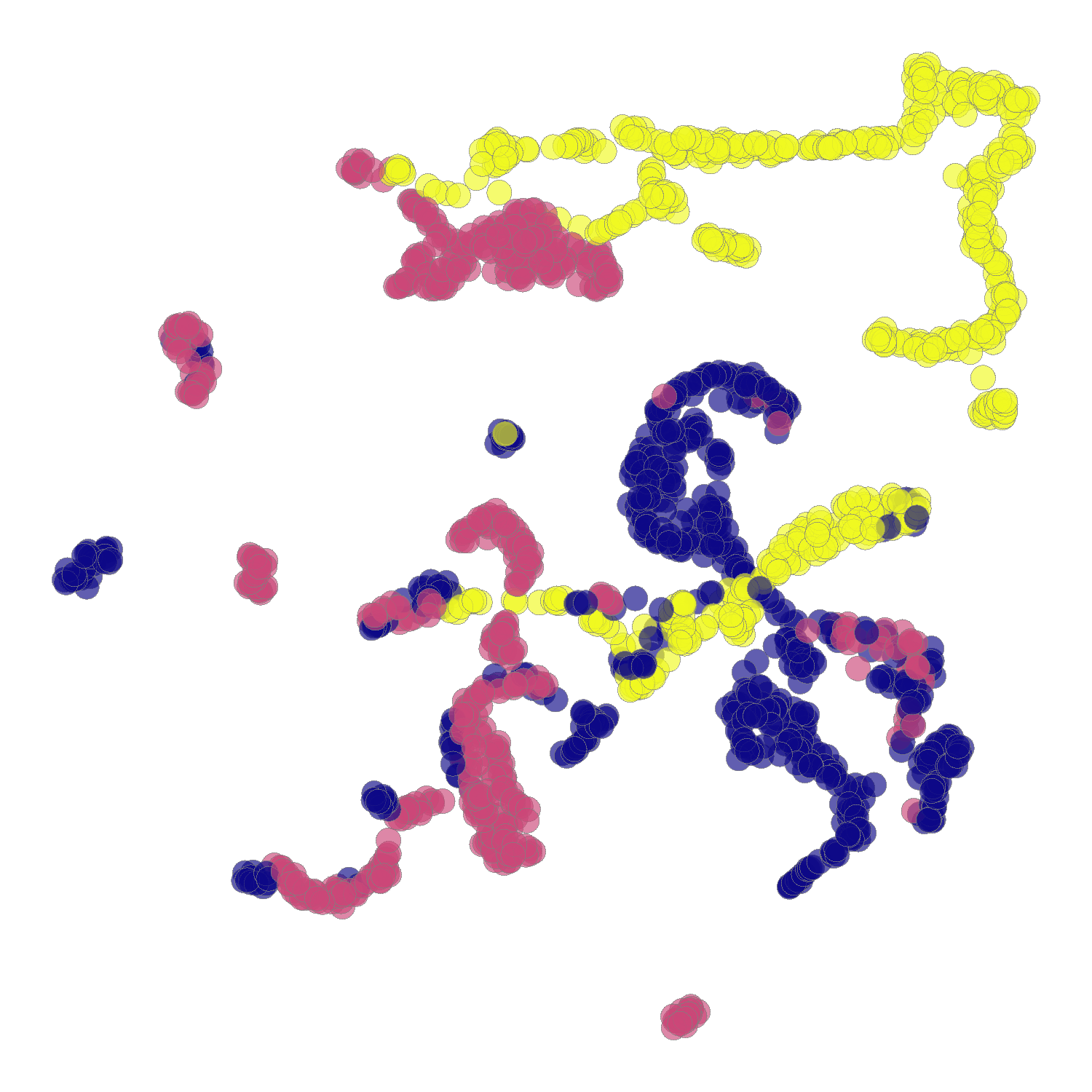}
        \caption{IAE pose}
    \end{subfigure}
    ~ 
    \begin{subfigure}[t]{0.15\textwidth}
        \centering
        \includegraphics[width=.98\textwidth]{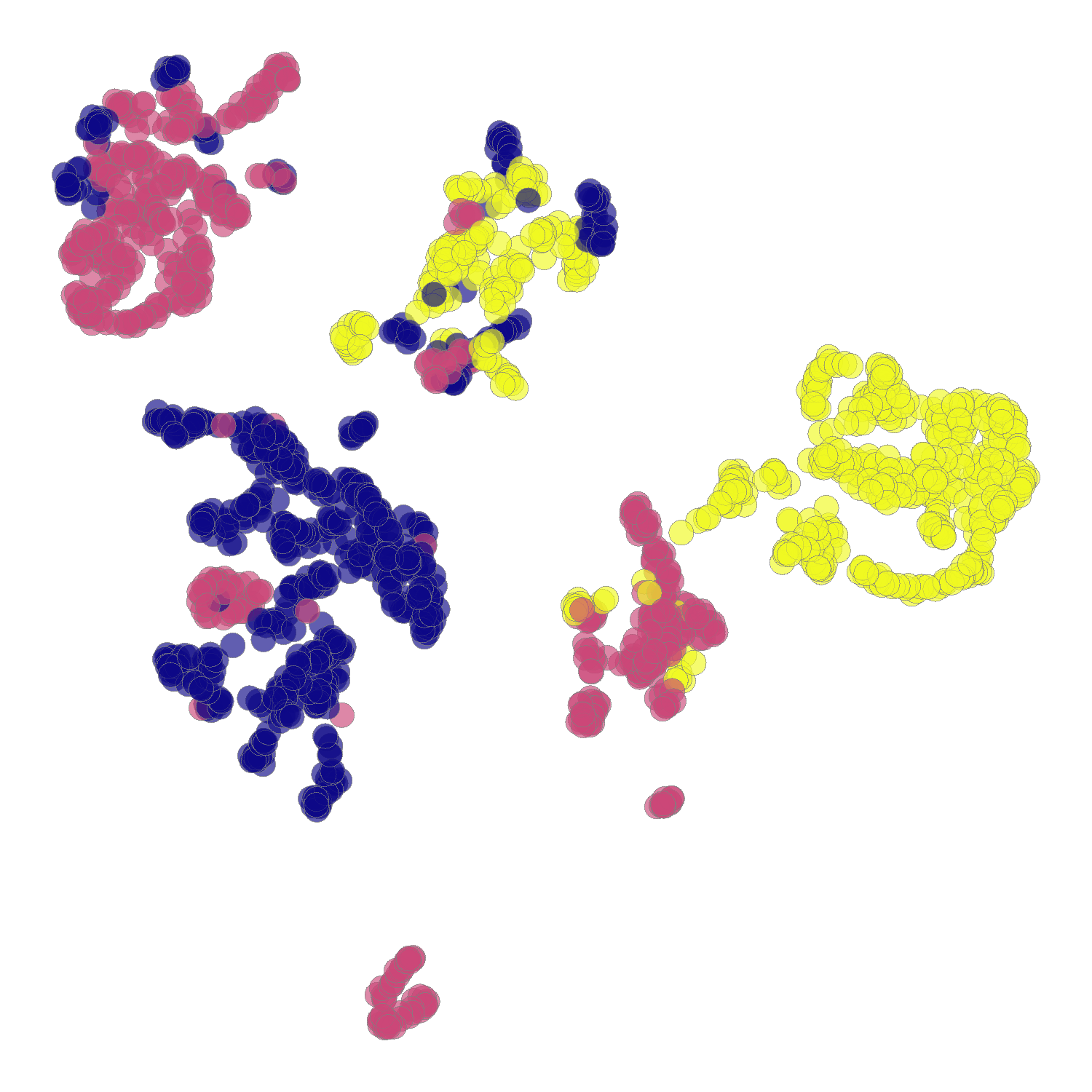}
        \caption{IAE merged}
    \end{subfigure}
    \begin{subfigure}[t]{0.15\textwidth}
        \centering
        \includegraphics[width=.98\textwidth]{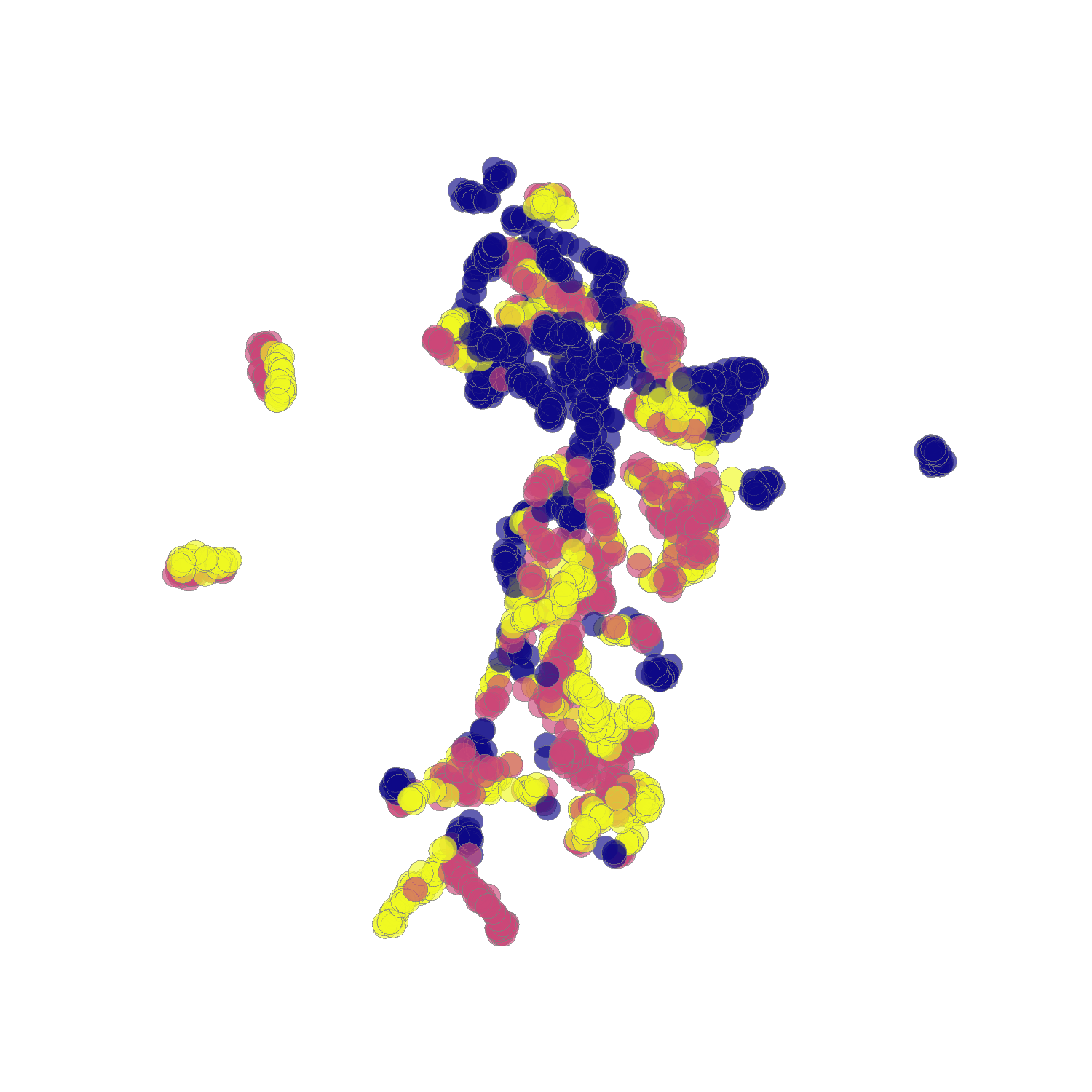}
        \caption{XAE face}
    \end{subfigure}%
    ~ 
    \begin{subfigure}[t]{0.15\textwidth}
        \centering
        \includegraphics[width=.98\textwidth]{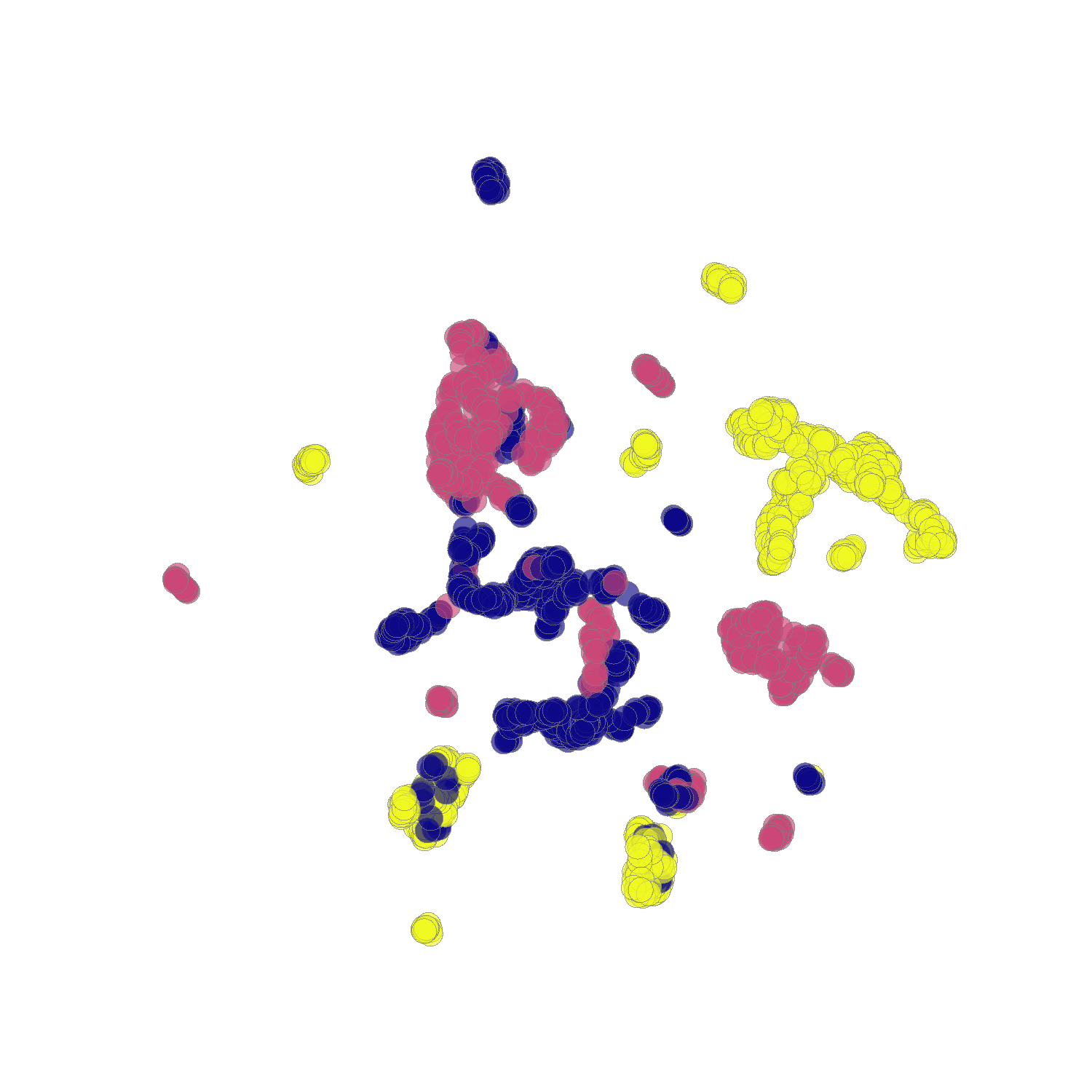}
        \caption{XAE pose}
    \end{subfigure}
    ~ 
    \begin{subfigure}[t]{0.15\textwidth}
        \centering
        \includegraphics[width=.98\textwidth]{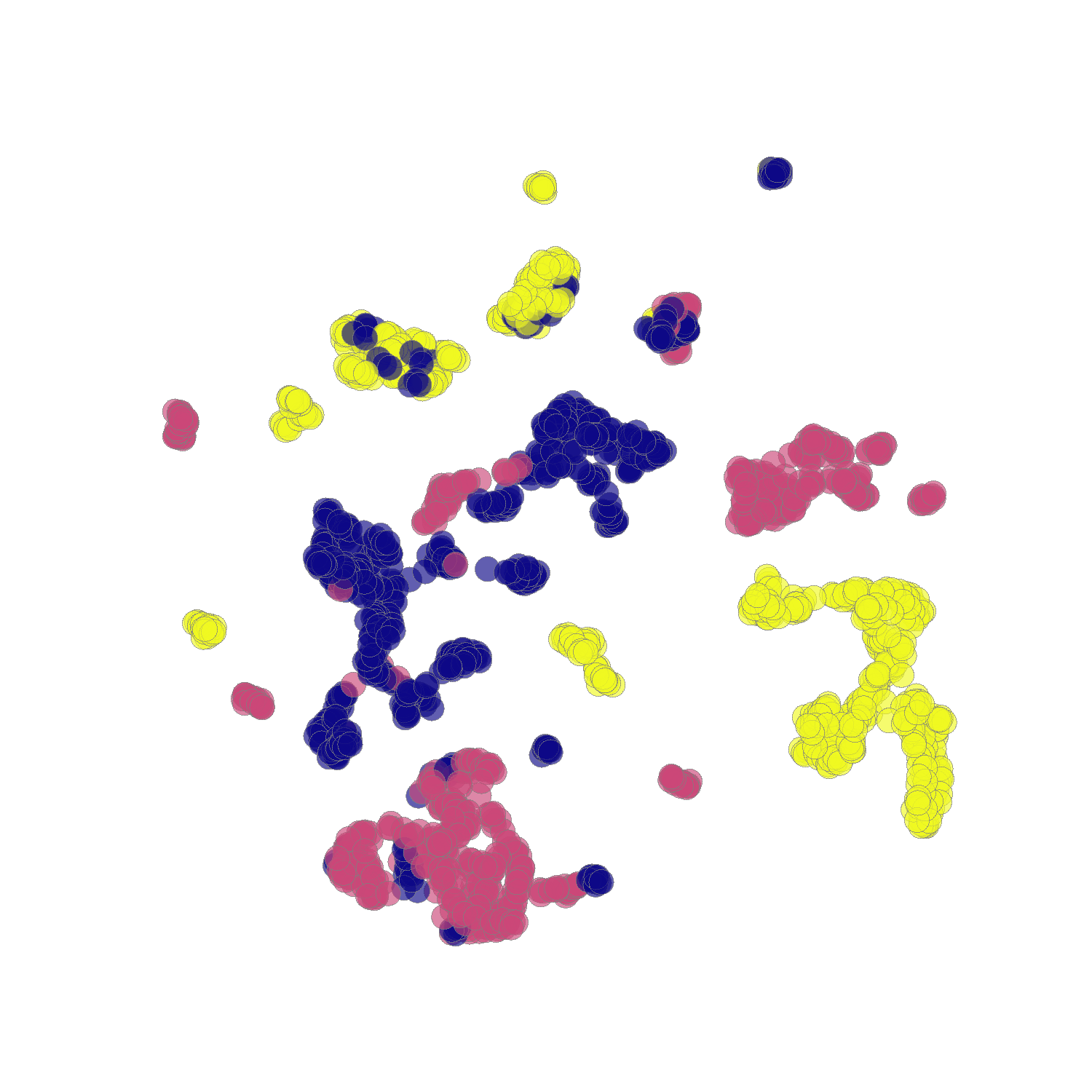}
        \caption{XAE merged}
    \end{subfigure}
    \caption{2-dimensional UMAP embeddings of face, pose, and their merged latent representations formed on the baseline model (top) vs. the explainable model (bottom). The individual colors demonstrate the labels of the individual inputs (blue=left, purple=right, yellow=robot).}
    \label{fig:umap}
\end{figure}

\begin{figure}[t]
	\centering
	\includegraphics[width=.60\textwidth]{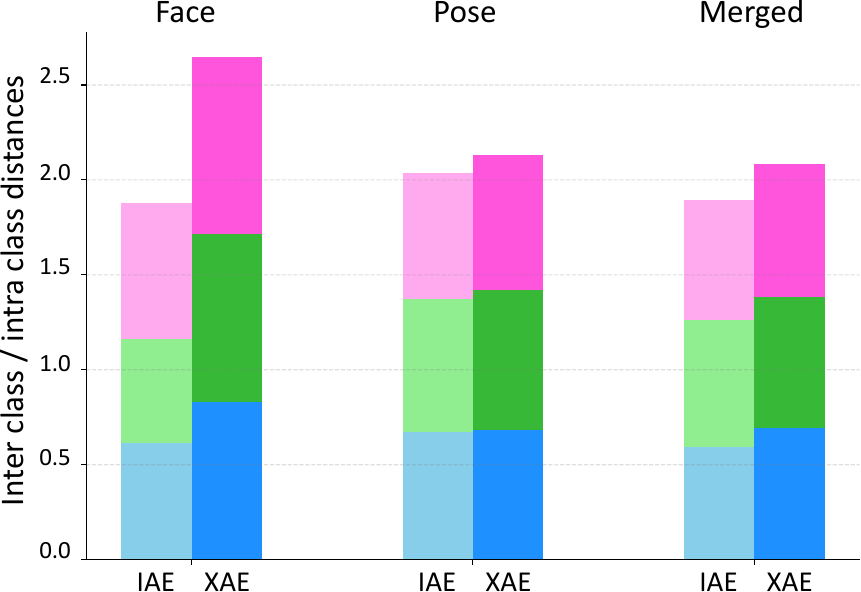}
	\caption{Development of ratio of inter-class to intra-class distances of face, pose, and their merged embeddings in IAE and XAE networks. Individual colors represent three (out of ten) randomly picked test sets used for the measurements.}
	\label{fig:distances}
\end{figure}

Our goal is to consistently compare the representations of data embeddings while keeping in mind that high-dimensional spaces often obstruct fair comparisons. Therefore, using UMAP, we transform the inner activations of the input into two-dimensional space.
We first empirically set the main UMAP parameters (minimal distance in the output space as 0.7, and the number of neighbors which contribute to the final approximation as 20) to produce embeddings that are neither too stretched nor completely mixed. Then, for a given test set, we collect all face, pose and their respective merged embeddings, and project them to 2D space separately. The resulting 2D projections are shown in Figure \ref{fig:umap}, where we project the embeddings in the IAE and XAE models.
There we see that the face embedding is not very clumped to each other in general, which means that the data is more variable. 
To numerically assess the data structure after the UMAP projections, we compute the ratio of inter-class to intra-class distances. In figure \ref{fig:distances}, we plot this ratio for the IAE and the XAE model, for three randomly picked test-sets. We see that the XAE model has a higher ratio in the case of the face embeddings. This can be understood as face embeddings being quite complex when processed by the XAE model.

\subsection{Evaluation of the modular architecture in real-time HRI.}

To measure the performance of our architecture in AE, we analyzed the results of the HRI evaluation study as reported in Section \ref{sec:methods:HRIstudy}.

\begin{figure*}[!htp]
	\centering
	\includegraphics[width=1\textwidth]{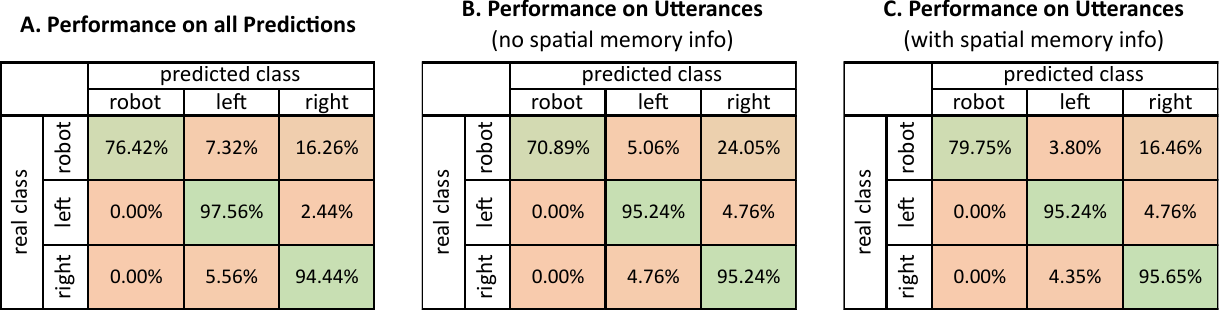}
	\caption{Confusion matrices representing the performances (recall) of the architecture with the deployed XAE model in the multi-party HRI evaluation study. Figure A represents the performance computed on all predictions of the XAE model (each utterance can lead to one or more predictions). Figure B represents the performance considering only the final estimate of XAE for each utterance. Figure C represents the performance on each utterance considering the correction made by the spatial memory module.}
	\label{fig:3confusion_matrices}
\end{figure*}

Considering all the predictions across all utterances for all experimental trials, the XAE model achieved a weighted F1-score of 85.56\%. Figure \ref{fig:3confusion_matrices}.A shows the confusion matrix, which reports the performance (recall) for each class. The figure shows an overall good accuracy with left and right classes performing better than the robot one (76.42\% for the robot class, 97.56\% for the left class, and 94.44\% for the right class). This demonstrates the validity and robustness of our approach also in real-time interaction scenarios.

However, the robot's behavior is ruled only by the final estimate of each utterance, so that the robot answers to the utterance in case of "robot" prediction" or turns left or right otherwise, in search of the predicted addressee. For this reason, it was essential to measure the performance of the XAE model only considering the final estimate. Figure \ref{fig:3confusion_matrices}.B shows the confusion matrix for the selected instance of the AE. The weighted F1-score computed on all utterances is 80.64\%. In this case, the accuracy of the robot class went slightly lower to 70.89\%, whereas the "left" and "right" classes maintained a similar score.

This estimate does not consider the spatial memory module, which is updated based on the people detected and remembered in the environment. Therefore, the very final AE from the architecture is the one with the applied corrections after considering the context from the spatial memory. Figure \ref{fig:3confusion_matrices}.C shows the accuracy of this final estimate. In the figure, an improvement in the estimation of the robot class is noticeable (79.75\%), which demonstrates a crucial role of the modular architecture in solving the AE task.

\begin{figure}[t]
	\centering
	\includegraphics[width=0.7\textwidth]{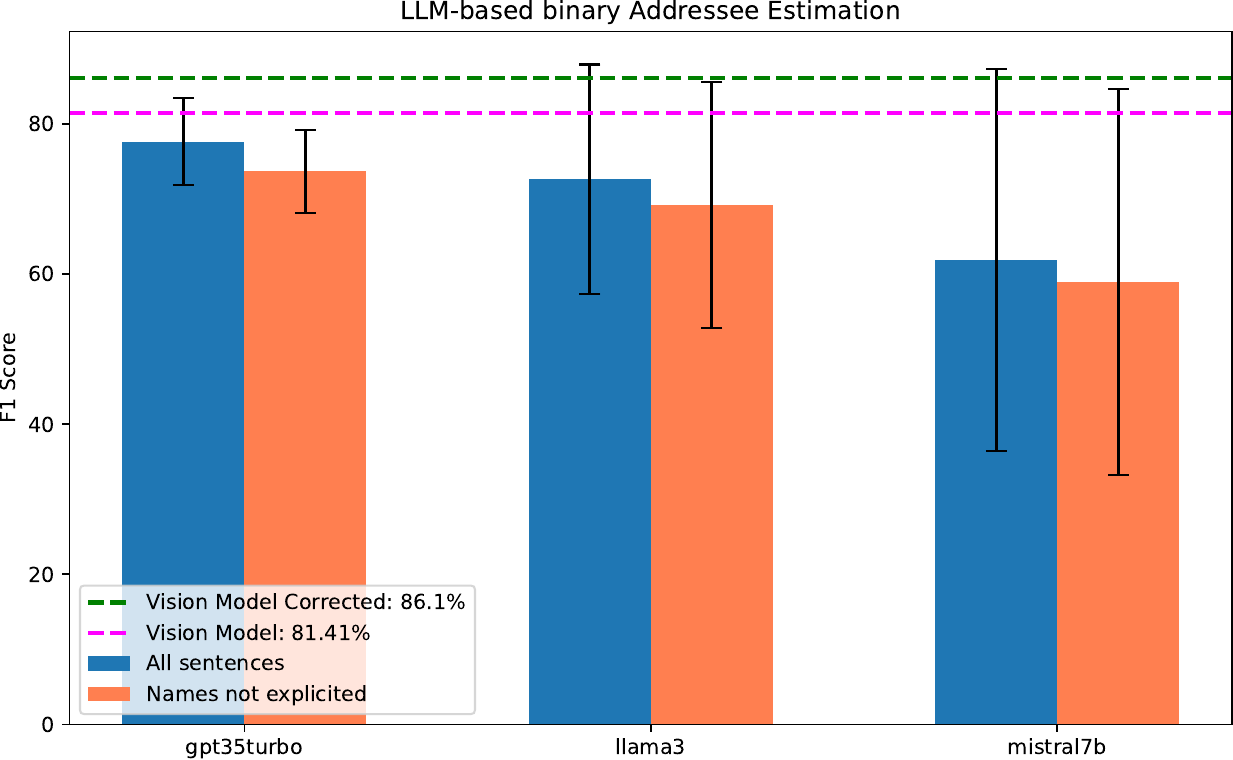}
	\caption{Average weighted F1 scores for addressee prediction using three different LLMs, based on conversations from the HRI evaluation study. For comparison, the corresponding metrics for the vision-based model are also reported, both with and without spatial memory correction.}
	\label{fig:EvaluationAE_llms}
\end{figure}

To assess whether the CV approach could be replaced by a linguistic approach with LLMs, we evaluated the performance of different families of LLMs in a binary task for predicting the correct addressee (ROBOT vs NOT-ROBOT) for conversations in the HRI evaluation study. We found that, on average, none of the three tested LLM outperformed the vision system (see Figure \ref{fig:EvaluationAE_llms}). In the same task, our XAE model achieved a weighted F1-score of 81.42\% on utterances, which increased up to 86.09\% with the correction of the spatial memory. On the other hand, the best LLM we tested is GPT-3.5-Turbo with an average F1-score of 77.6 $\pm$5.8\%, followed by llama3 (72.6$\pm$15.\%) and Mistral7b (61.8$\pm$25.4\%). Furthermore, when considering only sentences with no explicit reference to the addressee's identity, the LLMs performance drops further (F1-score of 73.6$\pm$5.5\% for GPT-3.5-Turbo, 69.2$\pm$16.4\% for Llama3 and 58.9$\pm$25.7\% for Mistral7b).

\subsection{User Evaluation of the Explainable and Transparent System}
\label{sec:online_study}
\begin{table}[htbp]
\centering
\caption{Statistics referring to participants' evaluation of the robot's Explainability and Transparency System for different modalities of communication (verbal, embodied, graphical).}
\begin{tabular}{l|ll|ll|ll}
                   & \multicolumn{2}{l|}{\textbf{Satisfaction}} & \multicolumn{2}{l|}{\textbf{Usefulness}} & \multicolumn{2}{l}{\textbf{Intrusiveness}} \\
\textit{Method}    & \textit{Mean}         & \textit{SD}        & \textit{Mean}        & \textit{SD}       & \textit{Mean}            & SD              \\ \hline
\textbf{verbal}    & 5.26                  & 0.89               & 5.04                 & 1.22              & 2.80                     & 1.66            \\
\textbf{embodied}  & 4.30                  & 1.37               & 4.38                 & 1.66              & 2.63                     & 1.61            \\
\textbf{graphical} & 5.03                  & 1.42               & 4.74                 & 1.74              & 3.28                     & 1.91           
\end{tabular}
\label{tab:questionnaires}
\end{table}

\begin{figure}[!ht]
	\centering
	\includegraphics[width=0.70\textwidth]{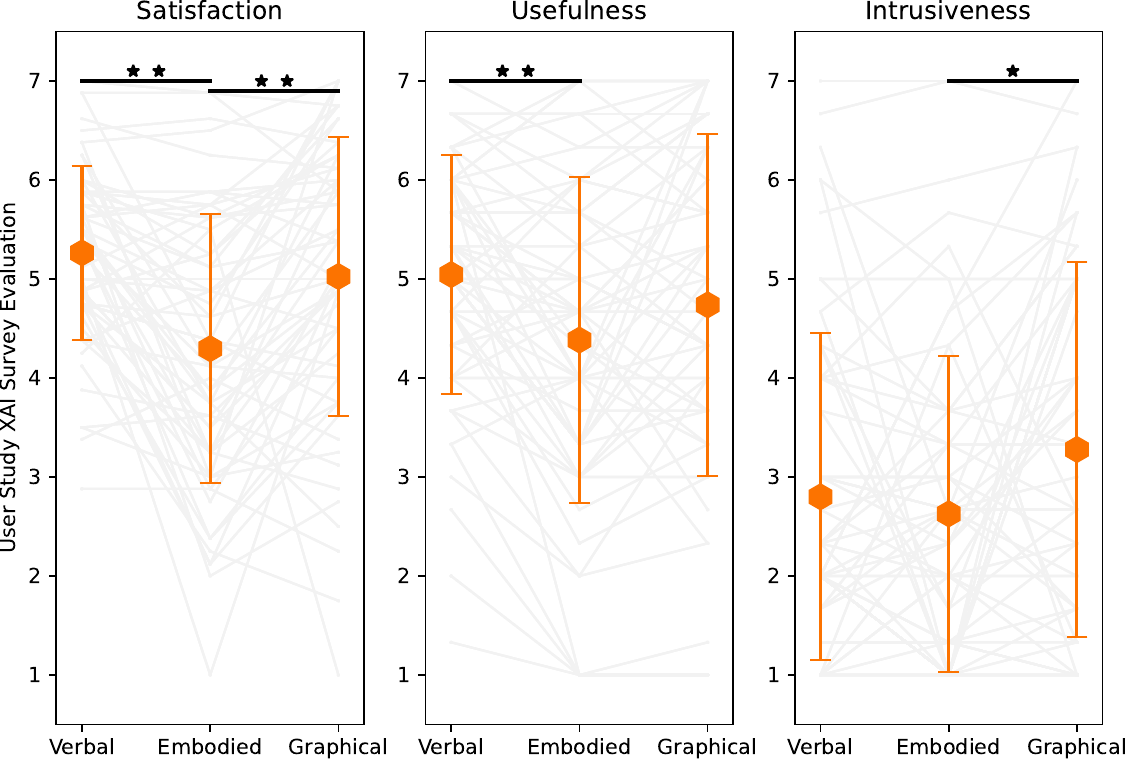}
	\caption{Plot of the Mixed Model on explanations' Satisfaction, Usefulness, and Intrusiveness. The orange hexagons represent the means of all modalities. Error bars are computed with SDs. Grey lines connect repeated measures for each participant, which in the Mixed Model are considered as a random effect. * refers to $p < 0.05$ and ** to $p < 0.001$.}
	\label{fig:mixed_model}
\end{figure}

Table~\ref{tab:questionnaires} reports the means and standard deviations of three questionnaire scales (Satisfaction, Usefulness, and Intrusiveness) grouped by the three explanation modalities (verbal, embodied, and graphical).
The Mixed Models analysis (see \ref{par:Analysis} for test description) revealed a significant Fixed effect of the modality on Satisfaction (F(2,118)=20.6, $p<0.001$), Usefulness (F(2,118)=9.24, $p<0.001$) and Intrusiveness (F(2,118)=4.62, p=0.012). After post-hoc pairwise comparisons with Bonferroni correction,  participants were found more satisfied with Verbal (M=5.26) and Graphical modalities than with the Embodied one (M=5.03) (Verb-Emb: B=0.968, t(118)=6.16, $p<.001$; Graph-Emb: B=0.729, t(118)=4.64, $p<0.001$), as shown in Figure~\ref{fig:mixed_model}. Concerning Usefulness, participants found more useful the Verbal (5.04) than the Embodied modality (4.38) (Verb-Emb: B=0.661, t(118)=4.30, $p<.001$) and more Intrusive the Graphical than the Embodied modality (Graph-Emb: B=0.650, t(118)=2.94, p=.012). The other comparisons were not found to be significant in a post-hoc test.

In the \hyperref[app:additional_analysis]{Appendix D}, Tables~\ref{tab:correlation_verbal}, \ref{tab:correlation_embodied}, \ref{tab:correlation_graphical} present all the results from Spearman Rank Correlations tests computed on the three parameters (Satisfaction, Usefulness and Intrusiveness). It is interesting to notice that for all modalities we found a strong positive correlation between Usefulness and Satisfaction. Moreover, a negative correlation was also found for all modalities between Satisfaction and Intrusiveness.

\begin{figure}[t]
	\centering
	\includegraphics[width=0.70\textwidth]{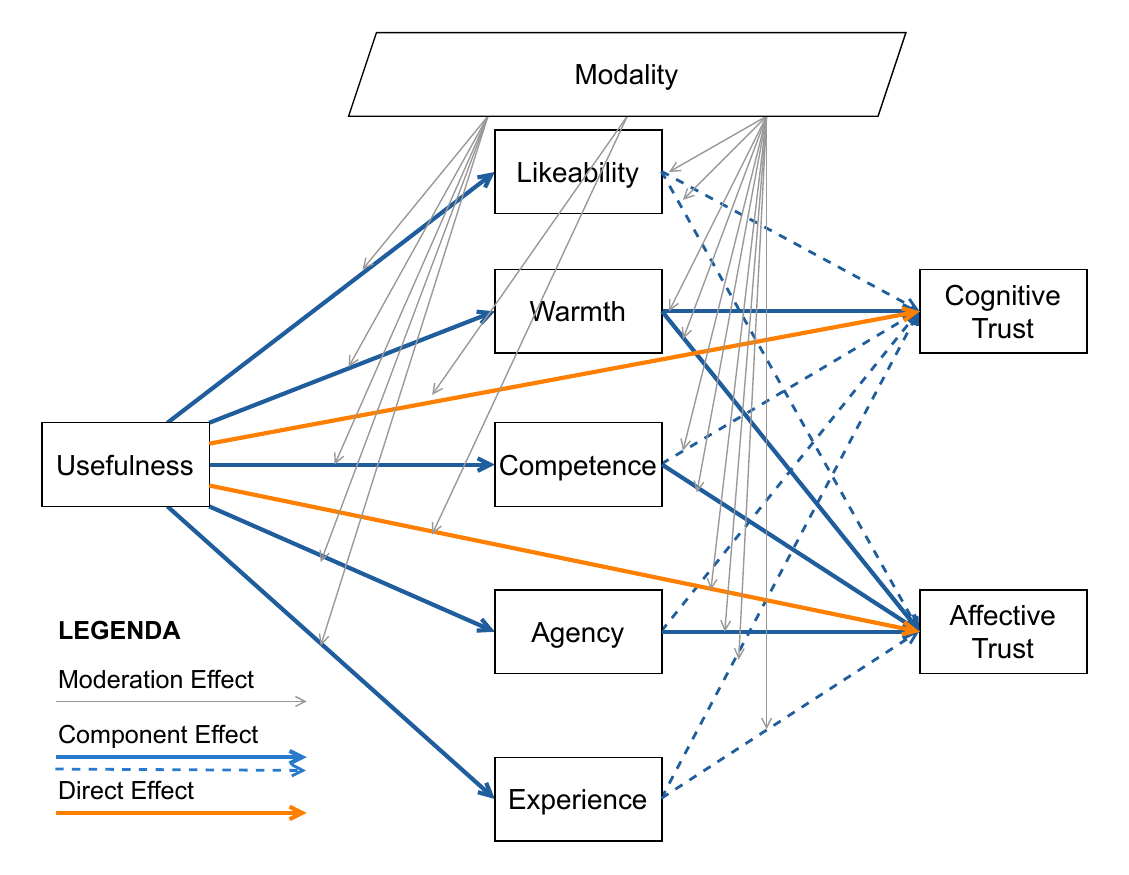}
	\caption{Path diagram of the Conditional Mediation analysis. The independent variable (or predictor) is the perceived Usefulness of explanations while the dependent variables are Affective Trust and Cognitive Trust. This relationship is mediated by participants' evaluation about several robots' qualities (Likeability, Warmth, Competence, Agency, Experience) and moderated by the three explanation Modalities (verbal, embodied, graphical). The solid orange and blue arrows represent, respectively, the significant direct and indirect (in the case of both components) effects of Usefulness on Trust.}
	\label{fig:pathdiagram}
\end{figure}

To further analyze participants' feedback provided in the online survey, we conducted a Conditional Mediation analysis, quantifying the extent to which the robot's qualities evaluated by participants (Competence, Likeability, Agency, Experience and Warmth) mediated the relationship between the perceived Usefulness of explanations and the trust toward the robot, and whether the explanation modalities (verbal, embodied, graphical) moderated such relationships. The path diagram in Figure ~\ref{fig:pathdiagram} displays the relationships and types of effects between the variables as assessed in the Conditional Mediation test. Overall (i.e., when averaging the effects associated with each explanation modality), we found a total effect (i.e., the combination of the direct and indirect effects) of Usefulness on both components of trust (on Affective Trust: $B = .42$, $SE = .05$, $p = \textless{}.001$, 95\% CI $= [.33, .51]$; on Cognitive Trust: $B = .31$, $SE = .04$, $p = \textless{}.001$, 95\% CI $= [.24, .38]$).
In regard to the relationship of Usefulness toward Affective Trust, we found a statistically significant effect both directly ($B = .10$, $SE = .05$, $p = .03$, 95\% CI $= [.01, .19]$) and indirectly. In this case, the test revealed a mediation of the robot's Agency ($B = .07$, $SE = .03$, $p = .02$, 95\% CI $= [.01, .12]$), Competence ($B = .15$, $SE = .04$, $p = \textless{}.001$, 95\% CI $= [.07, .24]$) and Warmth ($B = .09$, $SE = .04$, $p = .02$, 95\% CI $= [.01, .16]$). With respect to the relationship of Usefulness toward Cognitive Trust, the analysis revealed a direct effect ($B = .20$, $SE = .04$, $p = \textless{}.001$, 95\% CI $= [.11, .28]$) combined with an indirect effect mediated by the robot's Warmth  ($B = .10$, $SE = .04$, $p = .008$, 95\% CI $= [.03, .17]$). Moreover, as mentioned above, we also investigated the Conditional Mediation effect associated with each explanation modality, which we included in the \hyperref[app:additional_analysis]{Appendix D} (Tables \ref{tab:affective_trust} and \ref{tab:cognitive_trust}).

\section{Discussion}

This study is guided by the effort to put into action transparency and explainability techniques in a social robot with multi-party conversation abilities. To this aim, we designed and applied XAI solutions to a real-time human activity recognition model for the estimation of the addressee and implemented it in a modular robotic architecture for multi-party conversation together with other transparency solutions to show the underlying processing of the robot's behavior. 

\subsection{The explainability of our models}

Our ablation studies showed, that it is possible to train an addressee estimation network with a reasonable performance, regardless of which parts (M1, M2, M3) of the network are taken from the XAE architecture (explainable), and which from the IAE architecture (not explainable). Therefore, the choice of M1, M2, M3 can be used to regulate the explainability of the network. Thanks to the fact, that all tested networks achieved comparable performance scores, the choice of architecture depends only on the desired level of explainability. However, as explainable options are computationally more complex, it should be advised to use the simpler, not explainable version, in case the explanations are not needed at all.

When analyzing the heatmaps produced using Grad-CAM and those extracted from the XAE model, we saw that the XAE model usually produces clearer saliency maps that more correspond to the face location. Our hypothesis as to why the inherent explanations are better is due to the complex model architecture. Grad-CAM requires full gradient information computation for a correct estimate of the saliency map, but the gradients of a recurrent network can be relatively imprecise. 
Another benefit of computing the explanations on the XAE model is the computational efficiency. In addition to the inference, we only need a fraction of computational time since we only extract the already calculated attention scores. On the other hand, the Grad-CAM can be more tricky and it requires the gradients to be computed w.r.t. the last convolutional layer.

During our analysis of the hidden representations of face, pose, and merged information using the UMAP projection method, we found some differences between the XAE and IAE models. For example, the explainable model tends to create more separate point clusters (mainly of pose and merged embeddings), whereas the face embeddings are more entangled, and individual class instances are close to each other. Besides visual assessment of the produced projections, we also employed the computation of inter-class vs intra-class distances, that confirmed the observed variability of class separability, especially in the face representations.

Perhaps one of the greatest challenges is quantifiable explanation evaluation. To properly assess the level and usefulness of individual explainability methods, three main evaluation directions are often pursued: application-grounded, human-grounded, and functionally-grounded \citep{gilpin_explaining}. The first two involve human participants who provide meaningful feedback (i.e. in various user studies), which is later analyzed. However, the functionally-grounded evaluations are inherently difficult to achieve. There, the goal is to formulate a formal definition of explainability and a rigorous mathematical measure that produces quantitative outputs. For this reason, we have mostly focused on evaluating the quality of explanations by user studies.

\subsection{The real-time implementation in the architecture for multi-party conversation}

Built with a modular approach, our architecture was designed as the mean to connect our XAE model with the other modules necessary for the task of multi-party conversation: from Sound Detection to Spatial Memory and Speech Generation. At this stage, our architecture only missed a sound localization module, which the experimenter handled with the Wizard-of-Oz technique. Beyond that, the interaction flow was smooth and autonomous, as presented in the video.

Results from the HRI evaluation study highlighted a successful deployment of the XAE module in real-time interactions. The model was demonstrated to be robust, maintaining consistent performances with respect to the previous testing phase (see Section \ref{sec:AEmodel_performance}). In line with that, the model achieved better results for the left and right classes, confirming higher sensitivity in detecting poses and faces directed toward these two directions, whereas the accuracy dropped in the case of the speaker addressing the robot. This is particularly evident when computing the F1-score on the XAE final estimates, provided at the end of speakers’ utterances. Nevertheless, that is also the case in which the deployment of the XAE module in the modular architecture demonstrates to be fruitful. Thanks to the constantly updated spatial memory information, almost 30\% (7 on 24) of the wrong XAE predictions have been corrected, significantly improving the accuracy for the robot class and leading to an overall weighted F1-score of 86\%. More importantly, thanks to such corrections, the flow of interactions did not break, with the robot that was enabled to answer participants’ questions when addressed.

We decided to solve the AE problem with a CV approach because, as evidenced in previous literature \citep{Skantze2021}, meaningful information about the direction of the addressee are hidden in the speakers’ non-verbal behavior, specifically their gaze and body pose. To leverage such clues, we opted for processing both the image of the speakers’ faces and the coordinates vectors of their body joints. A different approach could be used, though. Information to estimate the addressee identity, albeit not their relative position in the space, might be contained also in the semantics of the utterance. It was therefore important to check whether such information is sufficient at least for a binary estimate (robot vs. somebody else) and, in this case, if the visual processing is redundant for this scope. To address this question, our analysis with different LLMs demonstrates that a purely linguistic approach for addressee prediction performs worse compared to the vision-based approach. Furthermore, as expected, performance decreases even further in cases where the addressee's name is not explicitly mentioned, which is a common scenario in real-world applications.
In other words, while semantic information serves as an important cue for addressing the problem of addressee estimation, it is often insufficient and not consistently reliable. A holistic approach that combines both types of observations and assigns greater weight to the one with higher confidence would be the optimal solution. To this end, we are actively developing a more advanced version of the proposed architecture that integrates behavioral and linguistic cues. This improvement is designed to provide greater robustness in handling complex and dynamic interaction scenarios, particularly those marked by high levels of uncertainty.

Another positive element about our approach is that, given the XAI techniques and system implemented in our architecture, the predictions of the robot are made less opaque, granting the user a greater awareness of the features and motivations underlying the robot’s behavior in real-time. That would not be the case with LLMs providing explanations that, besides being post-hoc, could also be verbose and reflect problems that LLMs still face, such as hallucination \citep{tonmoy_2024} and inaccuracy in reflecting the model's internal reasoning \citep{agarwal_2024}.

A modular design for the multi-party conversation architecture was preferred to an end-to-end approach for two reasons. First, multi-party conversation is a complex scenario involving various activities and multiple individuals interacting simultaneously. A modular approach offers greater controllability of the robot processes and behaviors in such unpredictable contexts.
Second, real-time multi-modal processing allows to exploit the synergy of different modules, enabling them to self-supervise each other and correct any erroneous estimates. 

As it is visible in the recorded multi-party interactions, the XAE model failed twice, but thanks to the modular approach, these errors could be amended, and the conversation resumed correctly.
For instance, thanks to Speech Understanding and Generation modules, when iCub is told the utterance was addressed to someone else, it apologizes for interrupting. 

The connection with spatial memory is another pivotal point. 
Thanks to this module, the final estimate of the addressee is significantly improved because it is not only based on multiple features coming from the robot's vision but also on continuously updated spatial-contextual information of the environment, following a more cognitively-inspired approach. 

Contextual information about people in the environment was used in previous works, but it was given as input to the neural network before the addressee estimation, without any possibility of updating it after the inference. Hence, to our knowledge, our architecture is the first that can 1) make corrections on the addressee identification based on additional exploration and spatial memory information and 2) discover new people and update the spatial memory based on the addressee estimation. For instance, in the first case, if the addressee is estimated to be on the speaker's left, but the robot doesn't detect anyone in that direction, it infers that the utterance was directed towards itself. On the other hand, in the second case, if the addressee is estimated at the speaker's left and the robot does not remember anybody in that direction, it turns its gaze to check over there and may detect new people.

\subsection{The users' evaluation}

The implementation of the XAI model in the robot's architecture for multi-party conversation aimed to enhance the comprehensibility of the robot's underlying processes. The system's opacity can be an obstacle to the perceived reliability of the robot (or any other artificial system) both for the users and its developers. Moreover, this issue becomes even more problematic when it comes to multi-faceted modular architectures, where several processes concur in executing a complex behavior \citep{wortham2017robot}. To reach an efficient and smooth interaction, developers and HRI designers need certainly to take care of the robot's performance, but also of their user-friendly intelligibility \citep{Sciutti_Humanizing}. 
It is to ensure this understandability, and hence reliability, that we designed our framework and system to provide real-time clarifications of the robot's behavior and the processes underlying its functioning. 

The act of explaining is a social mechanism: someone (the explainer) explains to someone else (the explainee) \citep{hilton1990conversational}. Recently, also the XAI community recognized and exploited this social component of the explainability problem, highlighting the explainees' needs within the explanation exchanges \citep{miller2019explanation}. The active role of the explainee has also been stressed by \cite{rohlfing2021explanation} in their co-constructive approach.

In our online user study, we presented multi-modal explanations to na\"ive explainees to assess their evaluation of the different modalities of the Explainability and Transparency System during some multi-party conversation scenarios. In this way, we aimed to quantify the explainability and transparency techniques deployed in the XAE model and in the whole robotic architecture from a user perspective, together with the users' perception of several features of the robot's appearance and behavior. 

We found all participants were more satisfied with the verbal and graphical explanations than with the embodied ones  
(i.e., the ones expressing the model's estimates and confidence via the robot's facial expressions). This result may be due to the embodied behavior design choices that some users could have found less intuitive and expressive than expected. Such a result is in line with the Usefulness metrics, where the verbal modality was found more useful than the embodied one. Interestingly, the graphical modality was found more Intrusive than the embodied one. The positive feature of non-verbal behavior is that it does not hinder the flow of interaction since it does not cause any interruption in verbal communication nor shift the users' focus of attention as the screen with graphical information may do. Nevertheless, this study shows that an accurate design of the robot's behavior is required because the implicit nature of non-verbal behavior may be an obstacle to the comprehension of the information conveyed.

On average, our Conditional Mediation analysis highlights the importance of the perceived usefulness of explanations to building trust towards robots, both in an affective and a cognitive sense. No relevant differences among explanation modalities were observed in the analysis of the Usefulness-Cognitive Trust relationship. Here, Usefulness maintains both a direct effect and an indirect effect mediated by Warmth, across all modalities. Conversely, in the analysis of the Usefulness-Affective Trust relationship, some differences became noticeable across modalities. This is especially evident for the verbal modality with respect to the others (embodied and graphical). With the verbal modality, the relation Usefulness-Affective Trust is not direct (it is so in the other modalities) but mediated (more strongly than the other modalities) by some robot's qualities: Agency, Competence, and Warmth. We can, therefore, hypothesize that while all modalities contributed to the building of trust, the contribution of the verbal explanations passed (more heavily) through the characterization of participants' perception of the robot.

Participants' satisfaction with the explanations and their perceived usefulness correlated for all the modalities meaning that participants' appreciation came in general from a utilitarian viewpoint: the more useful, the more satisfying. This result is coherent with the declared objective of the user study since participants had to exploit the robot's explanations to make sense of its behavior. All the modalities also showed a negative correlation between Satisfaction and Intrusiveness, showing that Intrusiveness should definitely be taken into account when designing an XAI system. 

Regarding the qualitative feedback provided by participants at the end of the survey, it is noteworthy that they never mentioned the robot's attention map in their open-ended responses. This suggests a reluctance to use that explanation modality, possibly due to difficulties in interpreting it.  

Overall, these results demonstrate that explainability represents a crucial capability for robots employed in social environments: not only their competence but also how and why they come to a decision matters. This is the reason why we do not need to arouse a blind trust towards robots but a justified one, based on the people's knowledge about robots' hidden processes, also exploiting the robots' embodiment \citep{matarese2021toward}. Diversifying the explanation modalities is one of the ways to follow when implementing XAI systems, as we showed in this study.

\subsection{Limitations and Future Improvements}
Implemented in the whole modular architecture, the explainable addressee estimation model we designed in this work offers verbal and non-verbal (embodied and graphical) explanations alongside the classification of the addressee. However, it is important to address its limitations. Given the surprisingly high sensitivity of the frequency and quality of the explanation on the hyper-parameters, a simple error optimization may not guarantee the model to produce meaningful explanations. Therefore, it would be useful to devise a training method that aligns with achieving good accuracy and generates reasonable and discriminative explanations.

In this work, we diverged from the majority of the previous approaches, which employed a simple binary classification, detecting only if the robot is addressed or not. The main reason was to provide the robot with extra information about the relative position of the addressee (in case it is not the robot), to enable smoother interactions. Although our 3-class classification approach is more fine-grained, in-depth comparisons of accuracy with previous methods are difficult to obtain. In theory, our approach could be replaced by performing multiple binary classifications for each output category and pooling the individual outputs to produce the final decision. However, due to the final pooling step, interpreting ensemble classifiers is much less straightforward. Therefore, we opted for a single multi-class classifier, which enabled us to explore multiple different explanation methods.

Although the XAE model has proven to be robust to a different setting, real-world deployment may require training with additional data because the model performance heavily relies on the quality and representativeness of the dataset, which may not always align with real-world scenarios. Thus, to improve its adaptability to new scenarios and robustness to different settings, an advisable solution would be implementing continual self-supervised learning to allow the robot to increment its knowledge while interacting with humans.Moreover, the current design of the modular architecture is not yet ready for exploitation in real-world scenarios. In our setting, the robot was limited in autonomy with regard to Speaker Identification abilities because the input of the sound source was provided by the experimenter. However, for a final version of the architecture, we are implementing a Sound Localization module to classify the direction of the upcoming sound source with respect to a robot-centric reference frame (e.g., from the right, front, or left of the robot). This will enable the activation of specific attention mechanisms, such as redirecting the robot's gaze toward the direction of the speakers' voice when they are outside its field of view.

Another limitation of the XAE deployment in real-time HRI lies in speaker diarization, a critical component for enabling seamless multi-party conversations. In our architecture, audio segmentation relied solely on silence intervals, leading to potential errors when participants spoke simultaneously or paused during their speech. To minimize confusion, participants were informed of these limitations and instructed to avoid speaking simultaneously and to leave sufficient pauses between their utterances, allowing the robot to accurately distinguish and process the speech segments. Moreover, the spatial memory version we employed was limited in granularity, making it insufficient for handling more complex interactions. To address these limitations, we plan to enhance the architecture by integrating a robust speaker diarization model and a more precise spatial memory system. Deploying the model in real-time also presented challenges related to network overload, particularly in maintaining a seamless operation of the modular architecture. As observed in the video, graphical explanations displayed on the TV screen occasionally flickered due to network saturation, which was more frequent during peak usage hours. The modular approach required extensive use of the local network to facilitate information exchange between modules, but this dependency occasionally led to performance bottlenecks. While this design ensures flexibility and scalability, we recognize the potential inconvenience it may cause for users in real-world scenarios and are exploring ways to optimize network usage to mitigate these issues.

Multi-party conversation is a complex challenge that requires robots to solve multiple tasks. We opted for a modular approach to design an architecture scalable to other human-activity recognition skills implemented as additional modules. Multi-modal emotion recognition, Speaker Diarization, Lips Movement Detection, and Voice Recognition are possible skills that can be added to our architecture in the future to increase the robot's awareness of the social dynamics in multi-party conversation. However, the modularity of our architecture and its deployment via YARP middleware also pose limitations. To be fully scalable, the modular architecture would require adapting the modules that are specifically tailored to the iCub robot (e.g., the iKinGazeCtrl module that controls the robot's oculomotor behavior and the other modules linked to it).

With respect to scalability, different is the case for the proposed AE models, which can be easily re-used in other robotic platforms, if provided with a visual input, which can be processed into the face and pose data streams. To further enhance the generalization capabilities of our modules, the visual input can be processed using a pre-trained ViT, and fine-tuned on the specific downstream task, even if the physical robot is different (and e.g., is looking at the scene from a different angle). Our architectural design also allows for a broader modality merge, where it is sufficient that we have a ubiquitous representation of each modality. This means that we can merge all sources to create the overall representation and still be able to explain the relative contributions of each input data streams.
Moreover, the attention modules used to generate explanations from the XAE model can be utilized in other scenarios as long as there are means to communicate the explanations to the users (i.e., audio output for verbal explanations and a screen for visual explanations).

The online user study was essential to assess how users evaluated the proposed Explanability and Transparency System. It confirmed the importance of leveraging multi-modality for communication in the HRI context but also revealed that the embodied modality needs to be carefully designed. While robots' non-verbal behavior can convey information without interrupting the flow of conversation, it may lead to ambiguous communication, making it less effective in supporting transparency. This was the case, for instance, of our iCub's facial expressions, which participants did not consider as much satisfying and useful as verbal communication. Future work should draw inspiration from this cross-disciplinary study by combining the development of robotic XAI solutions with an assessment of their actual evaluation by users.

\section{Conclusion}

The development of autonomous robots often aims at their deployment in social environments. Might they be hospitals, schools, restaurants, offices, or homes, robots are required to work safely and efficiently, two things that in human-populated contexts require social-like abilities to perceive the (social) world and act accordingly. To prove their reliability, autonomous systems must be transparent and explainable, two qualities that can increase the users' trust in robots if the former are put in a position to interpret the behavior of the latter.

This work first proposes an explainable machine learning model to solve the problem of addressee estimation, that is, figuring out to whom an interlocutor is speaking within a multi-party conversation. Next, we embedded such a model in a modular architecture to enable the iCub robot to actively participate in such complex interactions. Finally, we proposed a setting in which we assessed the feasibility of the overall architecture while collecting the impressions of users on the robot's explainability and transparency mechanisms.

Thanks to our attention-based approach, our model does not require any additional processing to provide explanations fast, while also increasing the level of transparency. The saliency map of the speaker's face, the relative importance of each input feature, and insights about which parts of the interaction affected the robot's estimate the most are acquired from our model and integrated with other techniques to clarify the opacity of the iCub robot's decision in a challenging scenario such as multi-party conversation.

When it comes to deploying deep neural networks in robots to predict human behavior and engage in social interaction, it is fundamental to adopt a unified approach. From neural network design to the implementation of computer vision algorithms into a modular architecture, ending with an exploratory user study, our work sought to do this by considering and integrating different perspectives to unveil the opacity of artificial systems designed to interact with us.

\section*{Acknowledgments}

The research leading to these results has received funding from the project titled TERAIS in the framework of the program Horizon-Widera-2021 of the European Union under the Grant agreement number 101079338.

I. Bečková and Š. Pócoš were also supported in part by The Slovak Research and Development Agency, project no. APVV-21-0105.

This Preprint has been prepared based on the following template: \url{https://github.com/kourgeorge/arxiv-style.git}

\bibliographystyle{apalike}
\bibliography{refs}

\newpage
\section*{Appendix A - List of abbreviations}
\label{app:abbreviations}

\renewcommand{\thefigure}{A.\arabic{figure}}
\renewcommand{\thetable}{A.\arabic{table}}
\renewcommand{\theequation}{A.\arabic{equation}}
\setcounter{figure}{0}
\setcounter{table}{0}
\setcounter{equation}{0}

\begin{table}[!h]
\centering
\arrayrulecolor[rgb]{0.2,0.2,0.2}
\begin{tabular}{ll}
\hline
\multicolumn{2}{l}{Abbreviations} \\ 
\hline
AE & Addressee Estimation \\
HRI & Human-Robot Interaction \\
IAE & Improved Addressee Estimation (model)\\
LLM & Large Language Models \\
MOT & Multi-Object Tracker \\
SOTA & State of the art \\
XAE & Explainable Addressee Estimation (model)\\
XAI & Explainable Artificial Intelligence\\
\hline
\end{tabular}
\arrayrulecolor{black}
\end{table}

\FloatBarrier
\section*{Appendix B - Design of the attention-based explainable AE model: Additional methods for merging modalities}
\label{app:additional_methods}
\renewcommand{\thefigure}{B.\arabic{figure}}
\renewcommand{\thetable}{B.\arabic{table}}
\renewcommand{\theequation}{B.\arabic{equation}}
\setcounter{figure}{0}
\setcounter{table}{0}
\setcounter{equation}{0}

 To merge the two data streams, we also tested some additional architectural concepts. However, these methods neither significantly surpassed the baseline accuracy nor brought clear explanations to the user.

 \subsection*{Modality merge using multi-dimensional attention}

 \noindent In this part, we are going to transform the two modalities of pose and image vector. We base it on the mechanism of multi-dimensional attention proposed in \cite{shen_dissan}, where the authors propose an attention scoring not only to weight individual value vectors but to give each of their elements a corresponding weight as well. For a given time frame $t$, using this procedure we are able to merge two modalities (outputs of the face and pose networks, $\boldsymbol{f}_t \in \mathbb{R}^{d_{\text{face}}}$ and $\boldsymbol{p}_t \in \mathbb{R}^{d_{\text{pose}}}$ respectively) into a single representation, which is then fed to the final, recurrent network. Using this method, the modalities are combined while taking into account their cross-relations. Our divergence from the traditional approach is that we do not combine multiple value vectors to a single representation, but we rather compute one value vector at a time and feed it to the RNN. The exact computation is following:

 \begin{equation}
 \boldsymbol{q}_t = \mathbf{W}^Q \boldsymbol{p}_t,\quad \boldsymbol{k}_t = \mathbf{W}^K \boldsymbol{f}_t,\quad \boldsymbol{v}_t = \mathbf{W}^V \boldsymbol{f}_t, 
 \end{equation}
 where the matrices, $\mathbf{W}^Q \in \mathbb{R}^{d_{q} \times d_{\text{pose}}}$, $\mathbf{W}^K \in \mathbb{R}^{d_{k} \times d_{\text{face}}}$ and $\mathbf{W}^V \in \mathbb{R}^{d_{v} \times d_{\text{face}}}$ are trainable parameters representing transformation of the corresponding vectors to query, key and value.

 The next step is to compute the importance weights for each of the elements of the value vector.
 \begin{equation}
     \boldsymbol{e}_t = \boldsymbol{W}^T_D \times act(\boldsymbol{W}_1 \times \boldsymbol{q}_t + \boldsymbol{W}_2 \times \boldsymbol{k}_t + \boldsymbol{b}),
 \end{equation}
 where $\boldsymbol{W}_1 \in \mathbb{R}^{d_{\text{inner}} \times d_{q}}$, $\boldsymbol{W}_2 \in \mathbb{R}^{d_{\text{inner}} \times d_{k}}$, $\boldsymbol{W}_D \in \mathbb{R}^{d_{\text{inner}} \times d_{v}}$ and $\boldsymbol{b} \in \mathbb{R}^{d_{\text{inner}}}$ are trainable parameters. 

 The input to the RNN network is a simple Hadamard product of $\boldsymbol{e}_t$ and $\boldsymbol{v}_t$. Thus, each element of the value vector has its corresponding weight. \\

 \subsection*{Modality merge using general attention}

 \noindent In this method we leverage the idea of general attention procedure \citep{att_overview}. After the input is processed, we end up with a representation of pose and face, $\boldsymbol{p_t}$ and $\boldsymbol{f_t}$, respectively. However in this scenario, $\boldsymbol{f_t} \in \mathbb{R}^{d_{\text{pose}}}$ is a vector but $\boldsymbol{p_t} \in \mathbb{R}^{n_{p} \times d_{\text{embed}}}$ is a series of vectors, each corresponding to a channel aggregation for a single pixel value after the convolutions.

 To continue with a general scheme of attention, we create our query, keys, and values as follows:
 \begin{equation}
 \boldsymbol{q}_t = \mathbf{W}^Q \boldsymbol{p}_t,\quad \boldsymbol{k}_{t,i} = \mathbf{W}^K \boldsymbol{f}_{t,i},\quad \boldsymbol{v}_{t,i} = \mathbf{W}^V \boldsymbol{f}_{t,i}, 
 \end{equation}
 where the matrices, $\mathbf{W}^Q \in \mathbb{R}^{d_{q} \times d_{\text{pose}}}$, $\mathbf{W}^K \in \mathbb{R}^{d_{k} \times d_{\text{face}}}$ and $\mathbf{W}^V \in \mathbb{R}^{d_{v} \times d_{\text{face}}}$ are trainable parameters representing transformation of the corresponding vectors to query, key and value.
 A scoring function is applied to return the corresponding weight to each of the components of the face representation:
 \begin{equation}
     \boldsymbol{e}_{t,i} = \boldsymbol{w}^T_D \times act(\boldsymbol{W}_1 \times \boldsymbol{q}_t + \boldsymbol{W}_2 \times \boldsymbol{k}_{t,i} + \boldsymbol{b}),
 \end{equation}
 where $\boldsymbol{W}_1 \in \mathbb{R}^{d_{\text{inner}} \times d_{q}}$, $\boldsymbol{W}_2 \in \mathbb{R}^{d_{\text{inner}} \times d_{k}}$, $\boldsymbol{W}_D \in \mathbb{R}^{d_{\text{inner}}}$ and $\boldsymbol{b} \in \mathbb{R}^{d_{\text{inner}}}$ are trainable parameters. 
 This is followed by an alignment step, where we normalize the contribution of each component of the face representation.
 \begin{equation}
     a_{t, i} = \rm{softmax}(e_{t,i}; \boldsymbol{e}).
 \end{equation}

 The final representation $\boldsymbol{r}$ of the input is a weighted sum of the face components.
 \begin{equation}
     \boldsymbol{r_t}=\sum_{i=1}^{n_{p}} a_{t,i} . \boldsymbol{v_{t,i}}.
 \end{equation}

\FloatBarrier
\section*{Appendix C - Deployment of the XAE in a modular architecture: Prompts for the LLM agent}
\label{app:prompts}
\renewcommand{\thefigure}{C.\arabic{figure}}
\renewcommand{\thetable}{C.\arabic{table}}
\renewcommand{\theequation}{C.\arabic{equation}}
\setcounter{figure}{0}
\setcounter{table}{0}
\setcounter{equation}{0}

\textit{Shopping Mall Context:}\\
``You are a service robot named iCub, working as an assistant in shopping mall.
You help customers who need information about shops inside the mall.
You give informations based on customers needs and preferences.
Be friendly and keep answers very short and concise!.''

\textit{Domestic Assistant Context:}\\
``You are a domestic assistant robot named iCub. You can do several tasks, like preparing drinks and food, and your role is to help accomplish this task when required. Be friendly and keep answers very short and concise!.''

\textit{Restaurant Context:}\\
``You are a service robot named iCub, working as a waiter for a restaurant. You help customers who need to order their food and drinks. Be friendly and keep answers very short and concise!.''

\textit{LLM-based AE Predictions Context:}\\
``You are a robot taking part in a conversation with one or more people. Given who is speaking (the Speaker) and what the speaker says (the Message), identify who is the sentence's addressee. Choose among only two options: [Myself, the ROBOT], [Another person, NOT THE ROBOT].
Answer in the format: Addressee: [option].''

%

 \section*{Appendix D - Online User Study: Additional Analysis}
 \label{app:additional_analysis}
\renewcommand{\thefigure}{D.\arabic{figure}}
\renewcommand{\thetable}{D.\arabic{table}}
\renewcommand{\theequation}{D.\arabic{equation}}
\setcounter{figure}{0}
\setcounter{table}{0}
\setcounter{equation}{0}
\subsection*{Analysis of relationships between the three explanation evaluation criteria via correlation matrices}

\FloatBarrier

\begin{table}[h!]
\caption{Correlation matrix in the verbal modality.} 
\label{tab:correlation_verbal}
\small
\centering
\arrayrulecolor[rgb]{0.2,0.2,0.2}
\begin{tabular}{llrlrlrl}
\multicolumn{8}{l}{Correlation Matrix} \\ 
\hline
\multicolumn{1}{c}{~} & \multicolumn{1}{c}{~} & \multicolumn{2}{c}{Satisfaction} & \multicolumn{2}{c}{Usefulness} & \multicolumn{2}{c}{Intrusiveness} \\ 
\hline
Satisfaction & Spearman's $\rho$ & — &  & ~ &  & ~ &  \\
~ & df & — &  & ~ &  & ~ &  \\
~ & p-value & — &  & ~ &  & ~ &  \\
Usefulness & Spearman's $\rho$ & .512 & *** & — &  & ~ &  \\
~ & df & 58 &  & — &  & ~ &  \\
~ & p-value & \textless{}.001 &  & — &  & ~ &  \\
Intrusiveness & Spearman's $\rho$ & -.293 & * & -.221 &  & — &  \\
~ & df & 58 &  & 58 &  & — &  \\
~ & p-value & .023 &  & .089 &  & — &  \\ 
\hline
\multicolumn{8}{l}{\textit{Note.} * p \textless{} .05, ** p \textless{} .01, *** p \textless{} .001} \\
\multicolumn{8}{l}{}
\end{tabular}
\arrayrulecolor{black}
\end{table}
\vspace{-0.5cm}

\begin{table}[h!]
\small
\caption{Correlation matrix in the embodied modality.}
\label{tab:correlation_embodied}
\centering
\arrayrulecolor[rgb]{0.2,0.2,0.2}
\begin{tabular}{llrlrlrl}
\multicolumn{8}{l}{Correlation Matrix} \\ 
\hline
\multicolumn{1}{c}{} & \multicolumn{1}{c}{~} & \multicolumn{2}{c}{Satisfaction} & \multicolumn{2}{c}{Usefulness} & \multicolumn{2}{c}{Intrusiveness} \\ 
\hline
Satisfaction & Spearman's $\rho$ & — &  & ~ &  & ~ &  \\
~ & df & — &  & ~ &  & ~ &  \\
~ & p-value & — &  & ~ &  & ~ &  \\
Usefulness & Spearman's $\rho$ & .770 & *** & — &  & ~ &  \\
~ & df & 58 &  & — &  & ~ &  \\
~ & p-value & \textless{}.001 &  & — &  & ~ &  \\
Intrusiveness & Spearman's $\rho$ & -.285 & * & -.262 & * & — &  \\
~ & df & 58 &  & 58 &  & — &  \\
~ & p-value & .027 &  & .043 &  & — &  \\ 
\hline
\multicolumn{8}{l}{\textit{Note.} * p \textless{} .05, ** p \textless{} .01, *** p \textless{} .001} \\
\multicolumn{8}{l}{}
\end{tabular}
\arrayrulecolor{black}
\end{table}
\vspace{-0.5cm}
\begin{table}[h!]
\footnotesize
\caption{Correlation matrix in the graphical modality.} 
\label{tab:correlation_graphical}
\centering
\arrayrulecolor[rgb]{0.2,0.2,0.2}
\begin{tabular}{llrlrlrl}
\multicolumn{8}{l}{Correlation Matrix} \\ 
\hline
\multicolumn{1}{c}{~} & \multicolumn{1}{c}{~} & \multicolumn{2}{c}{Satisfaction} & \multicolumn{2}{c}{Usefulness} & \multicolumn{2}{c}{Intrusiveness} \\ 
\hline
Satisfaction & Spearman's $\rho$ & — &  & ~ &  & ~ &  \\
~ & df & — &  & ~ &  & ~ &  \\
~ & p-value & — &  & ~ &  & ~ &  \\
Usefulness & Spearman's $\rho$ & .711 & *** & — &  & ~ &  \\
~ & df & 58 &  & — &  & ~ &  \\
~ & p-value & \textless{}.001 &  & — &  & ~ &  \\
Intrusiveness & Spearman's $\rho$ & -.473 & *** & -.376 & ** & — &  \\
~ & df & 58 &  & 58 &  & — &  \\
~ & p-value & \textless{}.001 &  & .003 &  & — &  \\ 
\hline
\multicolumn{8}{l}{\textit{Note.} * p \textless{} .05, ** p \textless{} .01, *** p \textless{} .001} \\
\multicolumn{8}{l}{}
\end{tabular}
\arrayrulecolor{black}
\end{table}

\FloatBarrier
\vspace{-0.5cm}
\subsection*{Gender effect on evaluation of explanation modalities}

Tables \ref{tab:gender_satisfaction} and \ref{tab:gender_usefulness} report the results of post-hoc tests from Mixed Model analysis on participants' evaluations of explanation modalities in terms of Satisfaction \ref{tab:gender_satisfaction} and Usefulness \ref{tab:gender_usefulness}.
While these models are structured as described in Section \ref{sec:online_study} (\textsection \hyperref[par:Analysis]{Questionnaires and Analysis}), we now included Gender as an additional factor. Therefore, Satisfaction, Usefulness, and Intrusiveness were used as dependent variables, while Modality and Gender were included as factors. The Intrusiveness model did not reveal any statistically significant results in post-hoc tests. Conversely, significant differences were found in the Satisfaction models when comparing both verbal and graphical modalities to embodied, and in the Usefulness model when comparing verbal to embodied. These differences were significant only within the female group; however, the main effect of Gender was not statistically significant in post-hoc tests across the three scales  (Satisfaction M.-F.: B=-.07, t(58)=-.21, \textit{p}=.83; Usefulness M.-F.: B=.621, t(58)=1.47, \textit{p}=.15; Intrusiveness M.-F.: B=-.06, t(58)=-.13, \textit{p}=.89). In conclusion, it remains unclear whether the modality differences observed only in the female group (in line with previous findings, see Section \ref{sec:AEmodel_analysis}) can be explained as a Gender effect or are instead due to the larger female sample size, potentially leading to limited statistical power for the male subgroup analysis. Future studies could further explore this potential impact as well as investigate whether participants' educational background (technical vs. non-technical) may also play a role.

\begin{table}[h!]
\small
\caption{Results from the post-hoc tests of the Mixed Model Analysis of Modality and Gender effects on Satisfaction}
\label{tab:gender_satisfaction}
\centering
\arrayrulecolor[rgb]{0.2,0.2,0.2}
\begin{tabular}{llrlrlrll}
\multicolumn{9}{l}{Post Hoc Comparisons - Modality (Mod.) * Gender (Gend.)}  \\ 
\hline
Mod.  & Gend. & Mod.  & Gend. & Diff. & SE   & t     & df  & \textit{p}                                                                                           \\ 
\hline
Graph. & M.    & Graph. & F.    & -.19  & 0.38 & -.49  & 111 & 1.00                                                                                        \\
Graph. & M.    & Non-v. & F.    & .65   & 0.38 & 1.69  & 111 & 1.00                                                                                        \\
Graph. & M.    & Verbal & F.    & -.44  & 0.38 & -1.16 & 111 & 1.00                                                                                        \\
\textbf{Non-v. }&\textbf{ F.}    & \textbf{Graph.} & \textbf{F.}    & \textbf{-.83} & \textbf{0.18} &\textbf{-4.64} &\textbf{116} & \textbf{\textless{}.001 }                                                                                       \\
Non-v. & M.    & Graph. & F.    & -.58  & 0.38 & -1.51 & 111 & 1.00                                                                                        \\
Non-v. & M.    & Graph. & M.    & -.39  & 0.33 & -1.20 & 116 & 1.00                                                                                        \\
Non-v. & M.    & Non-v. & F.    & .25   & 0.38 & 0.66  & 111 & 1.00                                                                                        \\
Non-v. & M.    & Verbal & F.    & -.84  & 0.38 & -2.18 & 111 & .471                                                                                        \\
Verbal & F.    & Graph. & F.    & .26   & 0.18 & 1.44  & 116 & 1.00                                                                                        \\
\textbf{Verbal} & \textbf{F.}    & \textbf{Non-v.} &\textbf{ F. }   & \textbf{1.09} & \textbf{0.18 }&\textbf{ 6.08}  & \textbf{116} & \textbf{\textless{}.001}                                                                                        \\
Verbal & M.    & Graph. & F.    & -.01  & 0.38 & -0.04 & 111 & 1.00                                                                                        \\
Verbal & M.    & Graph. & M.    & .17   & 0.33 & .53   & 116 & 1.00                                                                                        \\
Verbal & M.    & Non-v. & F.    & .82   & 0.38 & 2.13  & 111 & .525                                                                                        \\
Verbal & M.    & Non-v. & M.    & .56   & 0.33 & 1.73  & 116 & 1.00                                                                                        \\
Verbal & M.    & Verbal & F.    & -.27  & 0.38 & -.71  & 111 & 1.00                                                                                        \\ 
\hline                        
\end{tabular}
\arrayrulecolor{black}
\end{table}

\vspace{-0.5cm}
\begin{table}[h!]
\small
\caption{Results from the post-hoc tests of the Mixed Model Analysis of Modality and Gender effects on Usefulness}
\label{tab:gender_usefulness}
\centering
\arrayrulecolor[rgb]{0.2,0.2,0.2}
\begin{tabular}{llrlrlrll}
\multicolumn{9}{l}{Post Hoc Comparisons - Modality (Mod.) * Gender (Gend.)}                                                                                                \\ 
\hline
Mod.  & Gend. & Mod.  & Gend. & Diff. & SE  & t     & df   & p                                                                                           \\ 
\hline
Graph. & M.    & Graph. & F.    & .74   & .47 & 1.58  & 87.9 & 1.00                                                                                        \\
Graph. & M.    & Non-v. & F.    & 1.08  & .47 & 2.3   & 87.9 & .36                                                                                         \\
Graph. & M.    & Verbal & F.    & .37   & .47 & .78   & 87.9 & 1.00                                                                                        \\
Non-v. & F.    & Graph. & F.    & -0.34 & .18 & -1.93 & 116  & .85                                                                                         \\
Non-v. & M.    & Graph. & F.    & .34   & .47 & .72   & 87.9 & 1.00                                                                                        \\
Non-v. & M.    & Graph. & M.    & -0.4  & .32 & -1.26 & 116  & 1.00                                                                                        \\
Non-v. & M.    & Non-v. & F.    & .68   & .47 & 1.44  & 87.9 & 1.00                                                                                        \\
Non-v. & M.    & Verbal & F.    & -.04  & .47 & -.08  & 87.9 & .471                                                                                        \\
Verbal & F.    & Graph. & F.    & .38   & .18 & 2.13  & 116  & .524                                                                                        \\
\textbf{Verbal} & \textbf{F. }   & \textbf{Non-v. }&\textbf{ F.}    & \textbf{.72} &\textbf{.18} &\textbf{4.06}  & \textbf{116}  & \textbf{.001}                                                                                       \\
Verbal & M.    & Graph. & F.    & .82   & .47 & 1.73  & 87.9 & 1.00                                                                                        \\
Verbal & M.    & Graph. & M.    & .07   & .32 & .23   & 116  & 1.00                                                                                        \\
Verbal & M.    & Non-v. & F.    & 1.16  & .47 & 2.45  & 87.9 & .244                                                                                        \\
Verbal & M.    & Non-v. & M.    & .48   & .32 & 1.49  & 116  & 1.00                                                                                        \\
Verbal & M.    & Verbal & F.    & .44   & .47 & .93   & 87.9 & 1.00                                                                                        \\ 
\hline
\end{tabular}
\arrayrulecolor{black}
\end{table}

\subsection*{Insights on XAI user evaluation for each modality}

To understand how much explanations provided by the robot about its functioning contributed to increasing participants' trust in the robot, we inspected the relationship between the Usefulness of explanations and the two components of Trust (affective and cognitive) and assessed whether participants' perception of the robot's qualities played any role as a mediation factor. In particular, since we designed the robot's explanations to be provided via three different modalities (verbal, embodied and graphical) we verified whether the modalities moderated the abovementioned relationship. The following Tables report the results from the two Conditional Mediation tests designed to assess the abovementioned relationship. Affective and Cognitive Trust were, respectively, the two dependent variables, while Explanation Usefulness played the role of covariate, the five robot's qualities (Experience, Agency, Likeability, Competence and Warmth) were the mediators, and the explanation modality was the moderator.
\newpage

\begin{table*}[h!]
\caption{Results from Mediation Model: Usefulness effect on Affective Trust, mediated by robot's characteristics dimensions and moderated by XAI modality.}
\label{tab:affective_trust}
\centering
\scriptsize
\begin{tabular}{cclccccccc}
\multicolumn{10}{l}{Conditional Mediation} \\ 
\hline
Moderator levels & \multicolumn{1}{l}{} &  & \multicolumn{1}{l}{} & \multicolumn{1}{l}{} & \multicolumn{2}{c}{95\% C.I. (a)} & \multicolumn{1}{l}{} & \multicolumn{1}{l}{} & \multicolumn{1}{l}{} \\ 
\hline
Modality & Type & Effect & \multicolumn{1}{l}{Estimate} & \multicolumn{1}{l}{SE} & \multicolumn{1}{l}{Lower} & \multicolumn{1}{l}{Upper} & \multicolumn{1}{l}{$ \beta $} & \multicolumn{1}{l}{z} & \multicolumn{1}{l}{p} \\ 
\hline
Average & Indirect & $Usefulness \Rightarrow Experience  \Rightarrow Affective Trust $ & -.02 & .01 & -.04 & .01 & -.02 & -1.43 & .15 \\
\textbf{Average} &  & $\boldsymbol{Usefulness \Rightarrow Agency  \Rightarrow Affective Trust}$ & .07 & .03 & .01 & .12 & .09 & 2.42 & \textbf{.02} \\
Average &  & $Usefulness \Rightarrow Likeability  \Rightarrow Affective Trust$ & .03 & .02 & -.02 & .08 & .04 & 1.25 & .21 \\
Average &  & $\boldsymbol{Usefulness \Rightarrow Competence  \Rightarrow Affective Trust}$ & .15 & .04 & .07 & .24 & .21 & 3.66 & \textbf{\textless{}.001} \\
Average &  & $\boldsymbol{Usefulness \Rightarrow Warmth  \Rightarrow Affective Trust}$ & .09 & .04 & .01 & .16 & .12 & 2.28 & \textbf{.02} \\
Average & Component & $\boldsymbol{Usefulness \Rightarrow Experience}$ & .19 & .05 & .08 & .29 & .26 & 3.42 & \textbf{\textless{}.001} \\
Average &  & $Experience  \Rightarrow Affective Trust$ & -.09 & .06 & -.21 & .02 & -.09 & -1.58 & .11 \\
Average &  & $\boldsymbol{Usefulness \Rightarrow Agency}$ & .49 & .07 & .36 & .62 & .51 & 7.23 & \textbf{\textless{}.001} \\
Average &  & $\boldsymbol{Agency  \Rightarrow Affective Trust}$ & .14 & .05 & .03 & .24 & .19 & 2.57 & \textbf{.01} \\
Average &  & $\boldsymbol{Usefulness \Rightarrow Likeability}$ & .38 & .06 & .26 & .49 & .46 & 6.47 & \textbf{\textless{}.001} \\
Average &  & $Likeability  \Rightarrow Affective Trust$ & .08 & .06 & -.04 & .2 & .09 & 1.27 & .2 \\
Average &  & $\boldsymbol{Usefulness \Rightarrow Competence}$ & .48 & .05 & .38 & .58 & .61 & 9.45 & \textbf{\textless{}.001} \\
Average &  & $\boldsymbol{Competence  \Rightarrow Affective Trust}$ & .32 & .08 & .16 & .48 & .35 & 3.97 & \textbf{\textless{}.001} \\
Average &  & $\boldsymbol{Usefulness \Rightarrow Warmth}$ & .61 & .06 & .48 & .73 & .61 & 9.33 & \textbf{\textless{}.001} \\
Average &  & $\boldsymbol{Warmth  \Rightarrow Affective Trust}$ & .14 & .06 & .02 & .26 & .2 & 2.35 & \textbf{.02} \\
Average & Direct & $\boldsymbol{Usefulness \Rightarrow Affective Trust}$ & .1 & .05 & .01 & .19 & .14 & 2.18 & \textbf{.03} \\
Average & Total & $\boldsymbol{Usefulness \Rightarrow Affective Trust}$ & .42 & .05 & .33 & .51 & .59 & 8.87 & \textbf{\textless{}.001} \\
\hline
Verbal & Indirect & $Usefulness \Rightarrow Experience  \Rightarrow Affective Trust$ & -.01 & .01 & -.04 & .01 & -.02 & -.94 & .35 \\
Verbal &  & $\boldsymbol{Usefulness \Rightarrow Agency  \Rightarrow Affective Trust}$ & .09 & .04 & .01 & .17 & .12 & 2.22 & \textbf{.03} \\
Verbal &  & $Usefulness \Rightarrow Likeability  \Rightarrow Affective Trust$ & .03 & .03 & -.03 & .1 & .05 & 1.13 & .26 \\
Verbal &  & $\boldsymbol{Usefulness \Rightarrow Competence  \Rightarrow Affective Trust}$ & .24 & .07 & .11 & .36 & .32 & 3.6 & \textbf{\textless{}.001} \\
Verbal &  & $\boldsymbol{Usefulness \Rightarrow Warmth  \Rightarrow Affective Trust}$ & .13 & .05 & .03 & .23 & .17 & 2.47 & \textbf{.01} \\
Verbal & Component & $Usefulness \Rightarrow Experience$ & .18 & .12 & -.05 & .4 & .25 & 1.53 & .13 \\
Verbal &  & $Experience  \Rightarrow Affective Trust$ & -.07 & .06 & -.19 & .05 & -.07 & -1.19 & .24 \\
Verbal &  & $\boldsymbol{Usefulness \Rightarrow Agency}$ & .65 & .14 & .37 & .93 & .67 & 4.53 & \textbf{\textless{}.001} \\
Verbal &  & $\boldsymbol{Agency  \Rightarrow Affective Trust}$ & .14 & .05 & .03 & .24 & .18 & 2.54 & \textbf{.01} \\
Verbal &  & $\boldsymbol{Usefulness \Rightarrow Likeability}$ & .48 & .12 & .24 & .72 & .59 & 3.92 & \textbf{\textless{}.001} \\
Verbal &  & $Likeability  \Rightarrow Affective Trust$ & .07 & .06 & -.05 & .19 & .08 & 1.18 & .24 \\
Verbal &  & $\boldsymbol{Usefulness \Rightarrow Competence}$ & .66 & .11 & .45 & .87 & .85 & 6.17 & \textbf{\textless{}.001} \\
Verbal &  & $\boldsymbol{Competence  \Rightarrow Affective Trust}$ & .36 & .08 & .2 & .52 & .38 & 4.43 & \textbf{\textless{}.001} \\
Verbal &  & $\boldsymbol{Usefulness \Rightarrow Warmth}$ & .78 & .14 & .51 & 1.04 & .78 & 5.67 & \textbf{\textless{}.001} \\
Verbal &  & $\boldsymbol{Warmth  \Rightarrow Affective Trust}$ & .17 & .06 & .05 & .29 & .22 & 2.74 & \textbf{.006} \\
Verbal & Direct & $Usefulness \Rightarrow Affective Trust$ & .02 & .08 & -.14 & .19 & .03 & .29 & .77 \\
Verbal & Total & $\boldsymbol{Usefulness \Rightarrow Affective Trust}$ & .5 & .1 & .3 & .7 & .69 & 4.98 & \textbf{\textless{}.001} \\
\hline
Embodied & Indirect & $Usefulness \Rightarrow Experience  \Rightarrow Affective Trust$ & -.03 & .02 & -.07 & .00& -.04 & -1.71 & .09 \\
Embodied &  & $\boldsymbol{Usefulness \Rightarrow Agency  \Rightarrow Affective Trust}$ & .06 & .03 & .01 & .12 & .09 & 2.22 & \textbf{.03} \\
Embodied &  & $Usefulness \Rightarrow Likeability  \Rightarrow Affective Trust$ & .03 & .02 & -.01 & .07 & .04 & 1.38 & .17 \\
Embodied &  & $\boldsymbol{Usefulness \Rightarrow Competence  \Rightarrow Affective Trust}$ & .12 & .04 & .04 & .2 & .17 & 2.92 & \textbf{.003}\\
Embodied &  & $\boldsymbol{Usefulness \Rightarrow Warmth  \Rightarrow Affective Trust}$ & .07 & .04 & .00& .14 & .1 & 2.05 & \textbf{.04} \\
Embodied & Component & $Usefulness \Rightarrow Experience$ & .24 & .08 & .07 & .4 & .34 & 2.84 & \textbf{.004}\\
Embodied &  & $\boldsymbol{Experience  \Rightarrow Affective Trust}$ & -.13 & .06 & -.24 & -.01 & -.13 & -2.14 & \textbf{.03} \\
Embodied &  & $\boldsymbol{Usefulness \Rightarrow Agency}$ & .45 & .1 & .25 & .66 & .47 & 4.33 & \textbf{\textless{}.001} \\
Embodied &  & $\boldsymbol{Agency  \Rightarrow Affective Trus}t$ & .14 & .05 & .03 & .24 & .19 & 2.58 & \textbf{.01} \\
Embodied &  & $\boldsymbol{Usefulness \Rightarrow Likeability}$ & .33 & .09 & .16 & .51 & .41 & 3.69 & \textbf{\textless{}.001} \\
Embodied &  & $Likeability  \Rightarrow Affective Trust$ & .09 & .06 & -.03 & .21 & .1 & 1.49 & .14 \\
Embodied &  & $\boldsymbol{Usefulness \Rightarrow Competence}$ & .41 & .08 & .26 & .57 & .53 & 5.31 &\textbf{ \textless{}.001} \\
Embodied &  & $\boldsymbol{Competence  \Rightarrow Affective Trus}t$ & .28 & .08 & .12 & .44 & .31 & 3.5 & \textbf{\textless{}.001} \\
Embodied &  & $\boldsymbol{Usefulness \Rightarrow Warmth}$ & .55 & .1 & .35 & .74 & .55 & 5.45 & \textbf{\textless{}.001} \\
Embodied &  & $\boldsymbol{Warmth  \Rightarrow Affective Trust}$ & .13 & .06 & .02 & .25 & .19 & 2.21 & \textbf{.03} \\
Embodied & Direct & $\boldsymbol{Usefulness \Rightarrow Affective Trust}$ & .15 & .06 & .03 & .26 & .21 & 2.46 & \textbf{.01} \\
Embodied & Total & $\boldsymbol{Usefulness \Rightarrow Affective Trust}$ & .4 & .07 & .26 & .54 & .56 & 5.45 & \textbf{\textless{}.001} \\
\hline
Graphical & Indirect & $Usefulness \Rightarrow Experience  \Rightarrow Affective Trust$ & -.01 & .01 & -.03 & .01 & -.02 & -1.11 & .27 \\
Graphical &  & $\boldsymbol{Usefulness \Rightarrow Agency  \Rightarrow Affective Trust}$ & .05 & .02 & .00 & .1 & .07 & 2.12 & \textbf{.03} \\
Graphical &  & $Usefulness \Rightarrow Likeability  \Rightarrow Affective Trust$ & .02 & .02 & -.02 & .06 & .03 & 1.1 & .27 \\
Graphical &  & $\boldsymbol{Usefulness \Rightarrow Competence  \Rightarrow Affective Trust}$ & .12 & .04 & .04 & .19 & .16 & 3.07 & \textbf{.002} \\
Graphical &  & $Usefulness \Rightarrow Warmth  \Rightarrow Affective Trust$ & .06 & .03 & .00 & .13 & .09 & 1.93 & .05 \\
Graphical & Component & $Usefulness \Rightarrow Experience$ & .14 & .08 & -.01 & .3 & .2 & 1.79 & .07 \\
Graphical &  & $Experience  \Rightarrow Affective Trust$ & -.08 & .06 & -.2 & .03 & -.08 & -1.41 & .16 \\
Graphical &  & $\boldsymbol{Usefulness \Rightarrow Agency}$ & .37 & .1 & .17 & .57 & .38 & 3.7 & \textbf{\textless{}.001} \\
Graphical &  & $\boldsymbol{Agency  \Rightarrow Affective Trust}$ & .14 & .05 & .03 & .24 & .19 & 2.58 & \textbf{.01} \\
Graphical &  & $\boldsymbol{Usefulness \Rightarrow Likeability}$ & .32 & .09 & .15 & .49 & .39 & 3.69 & \textbf{\textless{}.001} \\
Graphical &  & $Likeability  \Rightarrow Affective Trust$ & .07 & .06 & -.05 & .19 & .08 & 1.15 & .25 \\
Graphical &  & $\boldsymbol{Usefulness \Rightarrow Competence}$ & .36 & .07 & .22 & .51 & .46 & 4.85 & \textbf{\textless{}.001} \\
Graphical &  & $\boldsymbol{Competence  \Rightarrow Affective Trust}$ & .32 & .08 & .16 & .48 & .35 & 3.97 & \textbf{\textless{}.001} \\
Graphical &  & $\boldsymbol{Usefulness \Rightarrow Warmth}$ & .5 & .1 & .31 & .68 & .5 & 5.18 & \textbf{\textless{}.001} \\
Graphical &  & $\boldsymbol{Warmth  \Rightarrow Affective Trust}$ & .13 & .06 & .01 & .25 & .18 & 2.09 & \textbf{.04} \\
Graphical & Direct & $\boldsymbol{Usefulness \Rightarrow Affective Trust}$ & .13 & .06 & .02 & .23 & .18 & 2.25 & \textbf{.03} \\
Graphical & Total & $\boldsymbol{Usefulness \Rightarrow Affective Trust}$ & .37 & .07 & .23 & .5 & .51 & 5.22 & \textbf{\textless{}.001} \\
\hline
\multicolumn{10}{l}{Note 1. Confidence intervals computed with method: Standard (Delta method). Note 2. Betas are completely standardized effect sizes} \\
\end{tabular}
\end{table*}
\newpage

\begin{table*}[h!]
\caption{Results from Mediation Model: Usefulness effect on Cognitive Trust, mediated by robot's characteristics dimensions and moderated by XAI modality.}
\label{tab:cognitive_trust}
\centering
\scriptsize
\begin{tabular}{cclccccccc}
\multicolumn{10}{l}{Conditional Mediation} \\ 
\hline
Moderator levels & \multicolumn{1}{l}{} &  & \multicolumn{1}{l}{} & \multicolumn{1}{l}{} & \multicolumn{2}{c}{95\% C.I. (a)} & \multicolumn{1}{l}{} & \multicolumn{1}{l}{} & \multicolumn{1}{l}{} \\ 
\hline
Modality & Type & Effect & \multicolumn{1}{l}{Estimate} & \multicolumn{1}{l}{SE} & \multicolumn{1}{l}{Lower} & \multicolumn{1}{l}{Upper} & \multicolumn{1}{l}{$ \beta $} & \multicolumn{1}{l}{z} & \multicolumn{1}{l}{p} \\ 
\hline
Average & Indirect & $Usefulness \Rightarrow Experience \Rightarrow
  Cognitive Trust$ & -.02 & .01 & -.04 & .01 & -.03 & -1.33 & .19 \\
Average &  & $Usefulness \Rightarrow Agency \Rightarrow Cognitive Trust$ & . & .03 & -.06 & .05 & -.01 & -.19 & .85 \\
Average &  & $Usefulness \Rightarrow Likeability \Rightarrow Cognitive Trust$ & .01 & .02 & -.04 & .05 & .01 & .32 & .75 \\
Average &  & $Usefulness \Rightarrow Competence \Rightarrow Cognitive Trust$ & .03 & .04 & -.05 & .1 & .05 & .75 & .45 \\
Average &  & $\boldsymbol{Usefulness \Rightarrow Warmth \Rightarrow Cognitive Trust}$ & .1 & .04 & .03 & .17 & .18 & 2.64 & \textbf{.008} \\
Average & Component & $\boldsymbol{Usefulness \Rightarrow Experience}$ & .19 & .05 & .08 & .29 & .26 & 3.42 & \textbf{\textless{}.001} \\
Average &  & $Experience \Rightarrow Cognitive Trust$ & -.08 & .06 & -.2 & .03 & -.11 & -1.44 & .15 \\
Average &  & $\boldsymbol{Usefulness \Rightarrow Agency}$ & .49 & .07 & .36 & .62 & .51 & 7.23 & \textbf{\textless{}.001} \\
Average &  & $Agency \Rightarrow Cognitive Trust$ & -.01 & .05 & -.11 & .09 & -.02 & -.19 & .85 \\
Average &  & $\boldsymbol{Usefulness \Rightarrow Likeability}$ & .38 & .06 & .26 & .49 & .46 & 6.47 & \textbf{\textless{}.001} \\
Average &  & $Likeability \Rightarrow Cognitive Trust$ & .02 & .06 & -.1 & .14 & .03 & .32 & .75 \\
Average &  & $\boldsymbol{Usefulness \Rightarrow Competence}$ & .48 & .05 & .38 & .58 & .61 & 9.45 & \textbf{\textless{}.001} \\
Average &  & $Competence \Rightarrow Cognitive Trust$ & .06 & .08 & -.1 & .21 & .08 & .75 & .45 \\
Average &  & $\boldsymbol{Usefulness \Rightarrow Warmth}$ & .61 & .06 & .48 & .73 & .61 & 9.33 & \textbf{\textless{}.001} \\
Average &  & $\boldsymbol{Warmth \Rightarrow Cognitive Trust}$ & .16 & .06 & .05 & .28 & .3 & 2.75 & \textbf{.006} \\
Average & Direct & $\boldsymbol{Usefulness \Rightarrow Cognitive Trust}$ & .2 & .04 & .11 & .28 & .36 & 4.48 & \textbf{\textless{}.001} \\
Average & Total & $\boldsymbol{Usefulness \Rightarrow Cognitive Trust}$ & .31 & .04 & .24 & .38 & .57 & 8.5 & \textbf{\textless{}.001} \\
\hline
Verbal & Indirect & $Usefulness \Rightarrow Experience \Rightarrow Cognitive Trust$ & -.01 & .01 & -.04 & .01 & -.02 & -.88 & .38 \\
Verbal &  & $Usefulness \Rightarrow Agency \Rightarrow Cognitive Trust$ & -.01 & .03 & -.08 & .06 & -.02 & -.31 & .76 \\
Verbal &  & $Usefulness \Rightarrow Likeability \Rightarrow Cognitive Trust$ & .01 & .03 & -.05 & .07 & .02 & .35 & .73 \\
Verbal &  & $Usefulness \Rightarrow Competence \Rightarrow Cognitive Trust$ & .03 & .05 & -.07 & .13 & .06 & .61 & .55 \\
Verbal &  & $\boldsymbol{Usefulness \Rightarrow Warmth \Rightarrow Cognitive Trust}$ & .13 & .05 & .03 & .23 & .25 & 2.59 & \textbf{.01} \\
Verbal & Component & $Usefulness \Rightarrow Experience$ & .18 & .12 & -.05 & .4 & .25 & 1.53 & .13 \\
Verbal &  & $Experience \Rightarrow Cognitive Trust$ & -.06 & .06 & -.18 & .05 & -.08 & -1.08 & .28 \\
Verbal &  & $\boldsymbol{Usefulness \Rightarrow Agency}$ & .65 & .14 & .37 & .93 & .67 & 4.53 & \textbf{\textless{}.001} \\
Verbal &  & $Agency \Rightarrow Cognitive Trust$ & -.02 & .05 & -.12 & .09 & -.03 & -.31 & .76 \\
Verbal &  & $\boldsymbol{Usefulness \Rightarrow Likeability}$ & .48 & .12 & .24 & .72 & .59 & 3.92 & \textbf{\textless{}.001} \\
Verbal &  & $Likeability \Rightarrow Cognitive Trust$ & .02 & .06 & -.1 & .14 & .03 & .35 & .73 \\
Verbal &  & $\boldsymbol{Usefulness \Rightarrow Competence}$ & .66 & .11 & .45 & .87 & .85 & 6.17 & \textbf{\textless{}.001} \\
Verbal &  & $Competence \Rightarrow Cognitive Trust$ & .05 & .08 & -.11 & .2 & .07 & .61 & .54 \\
Verbal &  & $\boldsymbol{Usefulness \Rightarrow Warmth}$ & .78 & .14 & .51 & 1.04 & .78 & 5.67 & \textbf{\textless{}.001} \\
Verbal &  & $\boldsymbol{Warmth \Rightarrow Cognitive Trust}$ & .17 & .06 & .06 & .29 & .32 & 2.91 & \textbf{.004} \\
Verbal & Direct & $\boldsymbol{Usefulness \Rightarrow Cognitive Trust}$ & .22 & .08 & .07 & .38 & .41 & 2.79 & \textbf{.005} \\
Verbal & Total & $\boldsymbol{Usefulness \Rightarrow Cognitive Trust}$ & .38 & .08 & .23 & .53 & .69 & 4.91 & \textbf{\textless{}.001} \\
\hline
Embodied & Indirect & $Usefulness \Rightarrow Experience \Rightarrow Cognitive Trust$ & -.03 & .02 & -.06 & .01 & -.05 & -1.63 & .1 \\
Embodied &  & $Usefulness \Rightarrow Agency \Rightarrow Cognitive Trust$ & . & .02 & -.05 & .04 & -.01 & -.13 & .89 \\
Embodied &  & $Usefulness \Rightarrow Likeability \Rightarrow Cognitive Trust$ & .01 & .02 & -.03 & .05 & .02 & .51 & .61 \\
Embodied &  & $Usefulness \Rightarrow Competence \Rightarrow Cognitive Trust$ & .02 & .03 & -.04 & .09 & .04 & .67 & .51 \\
Embodied &  & $\boldsymbol{Usefulness \Rightarrow Warmth \Rightarrow Cognitive Trust}$ & .09 & .04 & .02 & .17 & .17 & 2.57 & \textbf{.01} \\
Embodied & Component & $\boldsymbol{Usefulness \Rightarrow Experienc}e$ & .24 & .08 & .07 & .4 & .34 & 2.84 & \textbf{.004} \\
Embodied &  & $Experience \Rightarrow Cognitive Trust$ & -.12 & .06 & -.23 & . & -.15 & -2. & .05 \\
Embodied &  & $\boldsymbol{Usefulness \Rightarrow Agency}$ & .45 & .1 & .25 & .66 & .47 & 4.33 & \textbf{\textless{}.001} \\
Embodied &  & $Agency \Rightarrow Cognitive Trust$ & -.01 & .05 & -.11 & .1 & -.01 & -.13 & .89 \\
Embodied &  & $\boldsymbol{Usefulness \Rightarrow Likeability}$ & .33 & .09 & .16 & .51 & .41 & 3.69 & \textbf{\textless{}.001} \\
Embodied &  & $Likeability \Rightarrow Cognitive Trust$ & .03 & .06 & -.09 & .15 & .05 & .52 & .61 \\
Embodied &  & $\boldsymbol{Usefulness \Rightarrow Competence}$ & .41 & .08 & .26 & .57 & .53 & 5.31 & \textbf{\textless{}.001} \\
Embodied &  & $Competence \Rightarrow Cognitive Trust$ & .05 & .08 & -.1 & .21 & .07 & .67 & .5 \\
Embodied &  & $\boldsymbol{Usefulness \Rightarrow Warmth}$ & .55 & .1 & .35 & .74 & .55 & 5.45 & \textbf{\textless{}.001} \\
Embodied &  & $\boldsymbol{Warmth \Rightarrow Cognitive Trust}$ & .17 & .06 & .06 & .29 & .31 & 2.91 & \textbf{.004} \\
Embodied & Direct & $\boldsymbol{Usefulness \Rightarrow Cognitive Trust}$ & .17 & .06 & .06 & .29 & .31 & 2.94 & \textbf{.003} \\
Embodied & Total & $\boldsymbol{Usefulness \Rightarrow Cognitive Trust}$ & .27 & .06 & .16 & .38 & .49 & 4.73 & \textbf{\textless{}.001} \\
\hline
Graphical & Indirect & $Usefulness \Rightarrow Experience \Rightarrow Cognitive Trust$ & -.01 & .01 & -.03 & .01 & -.02 & -1.02 & .31 \\
Graphical &  & $Usefulness \Rightarrow Agency \Rightarrow Cognitive Trust$ & . & .02 & -.04 & .04 & . & -.13 & .9 \\
Graphical &  & $Usefulness \Rightarrow Likeability \Rightarrow Cognitive Trust$ & . & .02 & -.04 & .04 & . & .1 & .92 \\
Graphical &  & $Usefulness \Rightarrow Competence \Rightarrow Cognitive Trust$ & .03 & .03 & -.03 & .08 & .05 & .95 & .34 \\
Graphical &  & $\boldsymbol{Usefulness \Rightarrow Warmth \Rightarrow Cognitive Trust}$ & .07 & .03 & .01 & .13 & .13 & 2.21 & \textbf{.03}\\
Graphical & Component & $Usefulness \Rightarrow Experience$ & .14 & .08 & -.01 & .3 & .2 & 1.79 & .07 \\
Graphical &  & $Experience \Rightarrow Cognitive Trust$ & -.07 & .06 & -.19 & .04 & -.09 & -1.24 & .22 \\
Graphical &  & $\boldsymbol{Usefulness \Rightarrow Agency}$ & .37 & .1 & .17 & .57 & .38 & 3.7 &\textbf{ \textless{}.001} \\
Graphical &  & $Agency \Rightarrow Cognitive Trust$ & -.01 & .05 & -.11 & .1 & -.01 & -.13 & .89 \\
Graphical &  & $\boldsymbol{Usefulness \Rightarrow Likeability}$ & .32 & .09 & .15 & .49 & .39 & 3.69 & \textbf{\textless{}.001} \\
Graphical &  & $Likeability \Rightarrow Cognitive Trust$ & .01 & .06 & -.11 & .12 & .01 & .1 & .92 \\
Graphical &  & $\boldsymbol{Usefulness \Rightarrow Competence}$ & .36 & .07 & .22 & .51 & .46 & 4.85 & \textbf{\textless{}.001} \\
Graphical &  & $Competence \Rightarrow Cognitive Trust$ & .08 & .08 & -.08 & .23 & .11 & .97 & .33 \\
Graphical &  & $\boldsymbol{Usefulness \Rightarrow Warmth}$ & .49 & .1 & .31 & .68 & .5 & 5.18 & \textbf{\textless{}.001} \\
Graphical &  & $\boldsymbol{Warmth \Rightarrow Cognitive Trust}$ & .14 & .06 & .03 & .26 & .27 & 2.44 & \textbf{.02} \\
Graphical & Direct & $\boldsymbol{Usefulness \Rightarrow Cognitive Trust}$ & .2 & .05 & .09 & .31 & .37 & 3.66 & \textbf{\textless{}.001} \\
Graphical & Total & $\boldsymbol{Usefulness \Rightarrow Cognitive Trust}$ & .29 & .05 & .18 & .39 & .53 & 5.35 & \textbf{\textless{}.001} \\
\hline
\multicolumn{10}{l}{Note 1. Confidence intervals computed with method: Standard (Delta method). Note 2. Betas are completely standardized effect sizes} \\
\end{tabular}
\end{table*}
                                    
\end{document}